\documentclass[journal]{IEEEtran}
\usepackage{amsmath,amsfonts}
\usepackage{algorithmic}
\usepackage[switch]{lineno}
\usepackage{booktabs}
\usepackage{array}
\usepackage[font=normalsize,labelfont=sf,textfont=sf]{subcaption}
\usepackage{textcomp}
\usepackage{multirow}
\usepackage{multicol} 
\usepackage{tabularx} 
\usepackage{stfloats}
\usepackage{url}
\usepackage{verbatim}
\usepackage{orcidlink}
\usepackage{graphicx}
\usepackage{caption} 
\usepackage{subcaption} 
\def\BibTeX{{\rm B\kern-.05em{\sc i\kern-.025em b}\kern-.08em
    T\kern-.1667em\lower.7ex\hbox{E}\kern-.125emX}}
\usepackage{balance}
\usepackage{xcolor}
\usepackage[normalem]{ulem}

\usepackage{algorithm}
\usepackage{algorithmic}
\usepackage{booktabs}
\usepackage{tabularx}
\usepackage{multirow} 
\usepackage{multicol} 
\usepackage{amsmath}
\usepackage{amssymb}
\usepackage{amsthm}

\usepackage{xcolor}

\newcommand{\rev}[1]{\textcolor{black}{#1}}

\IEEEaftertitletext{%
\begin{center}
\footnotesize
\textit{This work has been submitted to the IEEE for possible publication. Copyright may be transferred without notice, after which this version may no longer be accessible.}
\end{center}

}
\begin{document}

\title{Burst Spiking Neural Networks}


 
\author{
Jiahong Zhang \orcidlink{0000-0002-5687-1839}, Sijun Shen \orcidlink{0009-0009-0654-6358}, Man Yao \orcidlink{0000-0002-0904-8524}, Han Xu \orcidlink{0000-0003-1411-3092}, Mingqiang Huang \orcidlink{0000-0002-7794-3985}, Yonghong Tian \orcidlink{0000-0002-2978-5935},~\IEEEmembership{Fellow,~IEEE}, \\
Bo Xu \orcidlink{0000-0002-1111-1529}, Guoqi Li \orcidlink{0000-0002-8994-431X}

\vspace{-3em}
\thanks{This work was partially supported by CAS Project for Young Scientists in Basic Research (YSBR-116), National Distinguished
Young Scholars (62325603), National Natural Science Foundation of China (62236009, U22A20103), Beijing Science and Technology Plan (Z241100004224011). (Corresponding author: Guoqi Li.)

Jiahong Zhang and Han Xu are with the Institute of Automation, Chinese Academy of Sciences, Beijing 100045, China, and the School of Artificial Intelligence, University of Chinese Academy of Sciences, Beijing 100049, China (e-mail: zhangjiahong2023@ia.ac.cn, xuhan2023@ia.ac.cn).

Sijun Shen is with State Key Laboratory of Media Convergence and Communication, Communication University of China, Beijing 100024, China (e-mail: sijun.shen@mails.cuc.edu.cn).

Mingqiang Huang is with School of Artificial Intelligence, Wuhan University, Wuhan, Hubei 430079, China (e-mail: mqhuang@whu.edu.cn).

Yonghong Tian is with Peng Cheng Laboratory, Shenzhen, Guangdong 518066, China, and the Institute for Artificial Intelligence, Peking University, Beijing 100871, China (e-mail: yhtian@pku.edu.cn).

Man Yao, Guoqi Li, and Bo Xu are with the Institute of Automation, Chinese Academy of Sciences, Beijing 100045, China (e-mail: man.yao@ia.ac.cn, guoqi.li@ia.ac.cn; xubo@ia.ac.cn).

}
}
\markboth{Submitted to IEEE Transactions on Pattern Analysis and Machine Intelligence}%
{Shell \MakeLowercase{\textit{et al.}}: A Sample Article Using IEEEtran.cls for IEEE Journals}

\IEEEpubid{}
\maketitle

\begin{abstract}

A central goal of current Spiking Neural Network (SNN) research is to improve their accuracy toward becoming low-power alternatives to Artificial Neural Networks (ANNs). \rev{This work further argues that realizing this ambition requires improving not only accuracy but also robustness, defined as the ability to maintain correct predictions under input perturbations. We identify two key issues in existing SNN methods that undermine robustness. First, binary spiking activations can produce large activation-state changes under small perturbations. Second, the lack of effective weight constraints makes network outputs more sensitive to input variations.} To this end, we propose Burst Spiking Neural Networks (BuSNNs), built upon Burst-enhanced Spiking Neurons (BSNs) and a Dynamic Weight Constraint (DWC) mechanism. \rev{BSNs incorporate burst firing to provide a graded spiking pattern. This spiking mechanism mitigates perturbation-induced transitions in activation states and thereby enhances robustness. DWC penalizes connection weights based on activation states, effectively reducing weight magnitudes and improving robustness while preserving accuracy. We provide theoretical analyses to support these robustness effects.} 
\rev{Experimental results further show that, on smaller-scale benchmarks such as CIFAR-10, BuSNNs outperform both SNN and ANN counterparts in accuracy and robustness. On large-scale ImageNet, BuSNN with the MS ResNet-34 backbone further improves top-1 accuracy and corruption robustness over the corresponding SNN baseline by 3.18\% and 2.66\%, respectively. Despite using spike-based activations, BuSNNs surpass 4-bit activation-quantized ANN baselines and approach 8-bit ANN baselines on ImageNet. They also preserve SNNs' low-power advantage with up to a 4.38$\times$ theoretical energy reduction over ANN counterparts.} This work studies the accuracy–robustness problem in SNNs, advancing their practical viability in robust and energy-efficient applications.

\end{abstract}


\section{Introduction}

Spiking Neural Networks (SNNs) have recently narrowed the accuracy gap with Artificial Neural Networks (ANNs) on vision benchmarks, driven by advances in deep learning~\cite{stbp,wu2019direct,han2020rmp,bu2022optimal,eshraghian2023training}. Their sparse and event-driven computation offers substantial gains in energy efficiency~\cite{yao2024spikenc,li2024brain}, making them attractive for deployment in resource-constrained environments. However, real-world scenarios often involve diverse perturbations and corruptions, making robustness a critical requirement. Realizing the advantages of SNNs in practical applications therefore depends crucially on improving both accuracy and robustness.

\begin{figure}[t]
    \centering
    \includegraphics[width=0.44\textwidth]{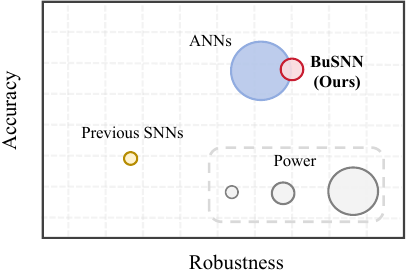}  
    \caption{
    Accuracy, robustness and power of SNNs and ANNs. Our method attains simultaneous improvements in accuracy and robustness and in some cases surpasses ANNs while preserving low power consumption.
    \color{black}{For instance, it surpasses its ANN counterparts on CIFAR-10 and CIFAR-100.}}
    
    \label{fig:acc_robust_trade1}
\end{figure}

\rev{
In this work, robustness is defined as the maximum variation in network outputs induced by bounded input perturbations. Smaller output variation indicates stronger robustness. In vision tasks, this property is typically evaluated under two settings: adversarial perturbations and common corruptions.
Adversarial perturbations are deliberately designed input changes that induce misclassification \cite{goodfellow2014explaining}, while common corruptions stem from conditions such as noise, blur, or weather effects~\cite{hendrycks2019benchmarking}.} 
Most existing SNN studies have primarily focused on improving adversarial robustness by exploiting spike coding, neural dynamics, and training strategies~\cite{el2021securing,xu2022securing,park2021noise,ZHANG2023164}. Recent approaches based on Poisson coding and adversarial training have achieved adversarial robustness exceeding that of ANNs~\cite{sharmin2020inherent,kim2022rate}. Poisson coding introduces stochastic spike trains to represent input intensities, effectively injecting noise that mitigates the impact of adversarial perturbations, but often at the cost of markedly reduced accuracy~\cite{wu2024rsc}.
In contrast, direct-coded SNNs deterministically convert input intensities into precise spike trains or rates \cite{plif,glif,clif}, achieving significantly higher accuracy and thus becoming the most widely used approaches \cite{psn,luo2024integer,yao2024spikedriven}.
Nevertheless, these models still fall short of ANNs in terms of adversarial robustness~\cite{kundu2021hire}.
Furthermore, our experiments reveal a clear limitation in the robustness of SNNs to common corruptions compared with ANNs. These observations highlight a persistent accuracy–robustness problem in SNNs, which prevents them from achieving ANN-level accuracy and robustness simultaneously, as illustrated in Fig.~\ref{fig:acc_robust_trade1}. 

This paper focuses on direct-coded SNNs, owing to their high accuracy and widespread use. We jointly examine their robustness under both adversarial attacks and common corruptions, aiming to achieve accuracy and robustness comparable to those of ANN counterparts. 
\rev{To this end, we propose Burst Spiking Neural Networks (BuSNNs), which focus on two critical factors: neural activation and connection weights.}

For neural activation, traditional perspectives suggest that single-spike neurons are inherently more robust to perturbations \cite{sharmin2019comprehensive,sharmin2020inherent,el2021securing}. 
\rev{However, binary spiking introduces severe quantization error, which limits representation capability and reduces accuracy~\cite{hu2023fast}.}
\rev{Our experiments in Section~\ref{bsnbinary} further show that binary spiking causes large perturbation-induced transitions in activation states.}
\rev{Recent studies have attempted to alleviate quantization error using multi-spike~\cite{luo2024integer,yao2025scaling} or burst-spike mechanisms~\cite{burst}. Motivated by these observations, we propose Burst-enhanced Spiking Neurons (BSNs). BSNs emit multiple spikes within a short temporal window and produce graded neuronal responses.} 
\rev{As a result, small perturbations either do not change the activation state or only induce bounded burst-count variations, reducing abrupt activation transitions and improving robustness.}

\rev{
Regarding connection weights, constraining weight magnitudes is known to improve network stability and robustness~\cite{cisse2017parseval,tsuzuku2018lipschitz}. For example, in a linear transformation $y=Wx+b$, smaller weight magnitudes reduce the sensitivity of outputs to input perturbations.
Biological synapses are also regulated in an activity-dependent manner under limited resources. Inspired by competitive and constrained synaptic plasticity~\cite{changeux1976selective,purves1980elimination,ooyen2001competition}, we propose a Dynamic Weight Constraint (DWC) mechanism within BuSNNs.
DWC applies different penalties to active and inactive connections according to neuronal activation states during training. Active connections are regularized to prevent excessive amplification. Inactive connections are suppressed to avoid accumulating large dormant weights. Together, these effects reduce perturbation amplification across layers and improve robustness.
}

Experimental results demonstrate that BuSNNs simultaneously enhance accuracy and robustness under both adversarial attacks and common corruptions. The key contributions are summarized as follows:

\begin{itemize}
    \item We provide a comprehensive evaluation of different SNN methods, benchmarking their adversarial and corruption robustness in a fair setting with shared network architectures and training strategies. The results demonstrate an accuracy–robustness gap for SNNs relative to ANNs.

    \item We identify two key robustness limitations in current SNNs and propose BuSNNs to address them. 
    \rev{BSNs reduce perturbation-induced state transitions through burst firing. DWC adaptively constrains weights according to activation states. Together, they improve robustness without sacrificing accuracy.}

    \item We validate our approach across multiple network architectures and datasets. Experimental results demonstrate consistent improvements in both clean and robust accuracy over existing SNNs. In particular, BuSNNs exceed other SNNs and ANN counterparts on CIFAR-10 and CIFAR-100. On ImageNet, they substantially narrow the performance gap between SNNs and ANNs, and surpass ANN counterparts under the ResNet-18 backbone.

\end{itemize}

\section{Related Work}

\rev{Accuracy and robustness are two central criteria for evaluating SNNs. Accuracy refers to classification performance on clean inputs, whereas robustness refers to the ability to maintain performance under perturbed inputs. In this paper, we discuss both adversarial robustness and corruption robustness.}

\subsection{Spiking Neural Networks}
SNNs are a class of biologically inspired neural networks that encode information with discrete spike events over time, mimicking the temporal dynamics and event-driven processing of biological neurons \cite{maass1997networks}. 
With the rise of neuromorphic computing and the demand for energy-efficient architectures, SNNs have attracted growing interest in various deep learning applications, including pattern recognition \cite{zhou2024direct,yao2024spike,SU2024106630,yao2024spikedriven,tclif,pmsn}, sensory processing \cite{wang2024spikevoice}, and motor control \cite{yu2023fault,yang2024firing}. Leveraging ANN-to-SNN conversion \cite{10502282,buoptimal,hu2023fast,wang2024universal,yuan2024trainable} and surrogate-gradient training \cite{stbp,wu2019direct,shrestha2018slayer}, SNNs have made significant performance advances.

SNN neuron designs are often inspired by neuroscience. Canonical spiking neuron models include Hodgkin–Huxley (H–H) \cite{hh}, Izhikevich \cite{izhikevich2003simple}, and the LIF \cite{lif} models. While the H–H model offers detailed biophysical realism and the Izhikevich model balances biological plausibility with efficiency, LIF-based neurons have become the prevailing standard for large-scale SNNs due to their computational simplicity. Extensions such as PLIF \cite{plif}, ALIF \cite{alif}, and GLIF \cite{glif} further increase expressiveness by incorporating adaptive thresholds, learnable membrane dynamics, and multi-timescale behaviors. Recent works have proposed more efficient and flexible neuron designs to reduce the temporal simulation burden inherent to SNNs. For instance, parallel spiking neuron models \cite{psn,li2024efficient,pmsn,mpsn} decouple spike computation across time steps, significantly accelerating inference while maintaining temporal resolution. These developments offer the potential for SNNs to scale effectively in practical tasks.

\begin{figure*}[t]
    \centering
    \includegraphics[width=0.96\textwidth]{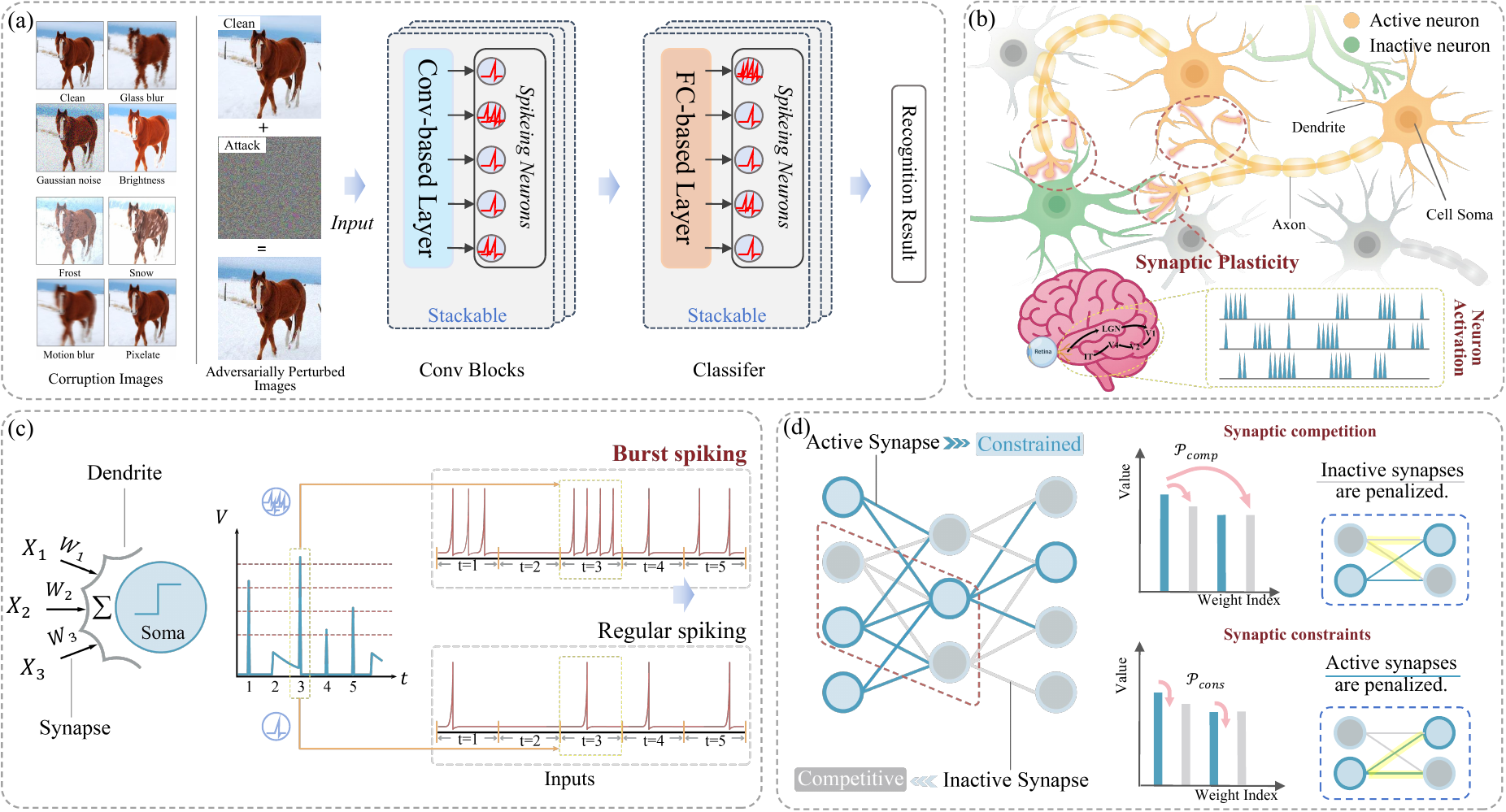}  
\caption{Overview of the proposed method. 
\rev{(a) The clean or corrupted image is processed by a deep neural network. Features are extracted by stackable convolutional (Conv) blocks with spiking neurons. Stackable fully connected (FC) layers then produce the recognition result.
(b)} In biological visual systems, neurons exhibit diverse activation patterns and form organized synaptic connections to respond effectively to external stimuli. \rev{(c)} We introduce the BSNs, which replaces conventional binary spiking to provide richer neuronal representations. \rev{(d)} Drawing inspiration from the competitive and constrained synaptic plasticity observed in biological neural systems, we propose a DWC mechanism that formalizes this principle to optimize connection weights from two complementary perspectives.}
    \label{fig:fig1}
\end{figure*}

\subsection{Robustness of Spiking Neural Networks}

Existing research on the robustness of SNNs has primarily focused on their resistance to adversarial attacks. They leveraged inherent SNN properties such as discrete spike encoding and non-continuous activation patterns, to obtain high adversarial robustness \cite{el2021securing,xu2022securing,park2021noise,sharmin2020inherent,kundu2021hire}. Various methods such as spike-based encoding schemes \cite{kim2022rate}, leaky membrane dynamics \cite{chowdhury2021towards}, architectural adaptations \cite{zhang2019fast,cheng2020lisnn}, and robustness-oriented training objectives \cite{kundu2021hire,ZHANG2023164,yang2023sibols} have been explored to further enhance this robustness.

However, deploying SNNs in real-world scenarios demands robustness of common corruptions, such as blur, variations in brightness and contrast \cite{hendrycks2019benchmarking}. Unlike adversarial noise, these corruptions arise from sensor imperfections or environmental fluctuations, rather than being deliberately crafted. In the ANN literature, extensive research has been conducted to enhance corruption robustness \cite{mixup,xie2020self,wang2023survey}. In contrast, research on corruption robustness in SNNs remains relatively limited. Recent efforts have indicated that existing SNNs often suffer substantial performance degradation on standard corruption benchmarks~\cite{mpsn}. 



\begin{table}[t]
\centering
\caption{{\color{black}Summary of representative SNN accuracy-robustness methods.}}
\label{tab:related_tradeoff}
{\color{black}
\footnotesize
\setlength{\tabcolsep}{3pt}
\begin{tabular}{p{0.34\columnwidth} p{0.27\columnwidth} p{0.31\columnwidth}}
\hline
Direction & Main Strength & Main Limitation \\
\hline

Poisson coding 
\cite{sharmin2020inherent,kim2022rate,wu2024rsc}
& Adversarial robustness
& Reduced clean accuracy \\

Direct-coded SNNs 
\cite{plif,glif,psn,mpsn}
& High clean accuracy
& Limited robustness \\

\hline
\end{tabular}
}
\end{table}

\subsection{Accuracy and Robustness Between SNNs and ANNs}


\rev{Robustness has been extensively studied in ANNs from multiple perspectives.
For common corruptions, benchmarks such as IMAGENET-C and KITTI-C provide systematic evaluation protocols~\cite{hendrycks2019benchmarking,dong2023benchmarking}.
Related benchmarks such as ImageNet-A and ObjectNet further evaluate natural distribution shifts and hard real-world examples~\cite{hendrycks2021natural,barbu2019objectnet}.
For adversarial perturbations, modern evaluations have exposed unreliable defenses and established stronger attack protocols~\cite{athalye2018obfuscated,croce2020reliable}.
Beyond evaluation, ANN studies have also developed robustness-oriented training and representation learning methods.
Representative examples include adversarial training~\cite{madry2017towards,Wang_2025_ICCV}, data augmentation~\cite{hendrycks2019augmix,ngnawe2023robustmix}, and knowledge distillation~\cite{xie2020self,zhang2020auxiliary,lu2025ciard}.
Some self-supervised learning studies further suggest that learned representations can improve both adversarial and corruption robustness~\cite{hendrycks2019using,zhang2022decoupled}.
Together, these ANN studies provide a broader context for evaluating SNN performance and robustness beyond spike-based models.}

It is often assumed that SNNs possess stronger robustness than ANNs due to their bio-inspired spike-based computation. This assumption has been mainly supported in the context of Poisson-coded SNNs. Previous studies show that SNNs using Poisson coding can outperform ANNs in adversarial robustness~\cite{sharmin2020inherent, kim2022rate, wu2024rsc}. \rev{Such robustness is commonly attributed to the stochastic and event-driven nature of Poisson spike generation, which can reduce the sensitivity of network responses to small adversarial perturbations. However, this encoding often leads to noticeably reduced clean accuracy.}

Direct-coded SNNs achieve significantly higher clean accuracy, making them widely used in modern SNN applications. However, their deterministic encoding reduces stochastic protection, which weakens the adversarial robustness observed in Poisson-coded SNNs~\cite{kundu2021hire}. Recent studies further indicate that SNNs still face substantial challenges under common corruptions~\cite{mpsn}. {These findings indicate that direct-coded SNNs still lack a clear mechanism for improving robustness while preserving their accuracy advantage.}

\rev{Overall, existing SNN robustness studies reveal an accuracy--robustness trade-off, as summarized in Table~\ref{tab:related_tradeoff}. Poisson-coded SNNs benefit from stochastic spike generation, but they often sacrifice clean accuracy. Direct-coded SNNs improve recognition performance, but their robustness remains limited. This trade-off reveals a key limitation of existing SNN methods. To address this gap, BuSNN improves direct-coded SNNs from two complementary aspects. BSNs enhance spike-based activation representation through bounded burst responses. DWC constrains synaptic weights according to activation-dependent importance. Together, these designs allow BuSNN to improve clean accuracy, adversarial robustness, and corruption robustness simultaneously.}

 \section{Burst Spiking Neural Networks}
 \label{section:method}

This work proposes BuSNNs to enhance SNN robustness from two complementary perspectives: neural activation and connection weights. \rev{Fig.~\ref{fig:fig1}(a) illustrates the {processing} pipeline, where clean, corrupted, or adversarially perturbed images are processed by a network to generate recognition results.} On the activation side, we introduce BSNs, which employ burst firing to produce graded activations and reduce perturbation-induced state transitions. On the weight side, we design a DWC mechanism that adaptively penalizes connection weights based on BSN activation states, effectively constraining weight magnitudes and lowering sensitivity to input perturbations. Both BSNs and DWC are generic modules that can be seamlessly integrated into existing SNN architectures without modifying their overall design.

\subsection{\rev{Unified Output-Stability View of Robustness}}

\rev{We define robustness from the perspective of output stability. Let \(f_\theta(x)\in\mathbb{R}^K\) denote the output logits of a network with parameters \(\theta\). For a clean input \(x\) and its perturbed counterpart \(x'\), we measure the perturbation-induced output variation as:}
{\color{black}\begin{equation}
\Delta_\theta(x,x') = \|f_\theta(x') - f_\theta(x)\|_\infty .
\label{eq:delta_output}
\end{equation}}
\rev{Adversarial and corruption robustness can then be described by the same quantity under different perturbation rules. For adversarial perturbations bounded in an \(\ell_p\) ball, we consider the worst-case output variation:}
{\color{black}\begin{equation}
\mathcal{R}_{\mathrm{adv}}(x;\theta,\epsilon)
= \sup_{\|x'-x\|_p \le \epsilon} \Delta_\theta(x,x') .
\label{eq:radv}
\end{equation}}
\rev{For common corruptions sampled from a corruption family \(\mathcal{C}\), we consider the expected output variation:}
{\color{black}\begin{equation}
\mathcal{R}_{\mathrm{corr}}(x;\theta)
= \mathbb{E}_{x'\sim \mathcal{C}(\cdot|x)}
\Delta_\theta(x,x') .
\label{eq:rcorr}
\end{equation}}

\rev{Thus, both robustness notions reduce to controlling the output variation \(\Delta_\theta(x,x')\). This formulation also connects to classification stability. \rev{For a correctly classified clean input \(x\), let \(y\in\{1,\ldots,K\}\) denote its ground-truth class, and let \(f_\theta(x)_j\) denote the logit corresponding to class \(j\). We define the clean logit margin as \(m_\theta(x)=f_\theta(x)_y-\max_{j\neq y}f_\theta(x)_j\).
This margin measures the separation between the ground-truth logit and the largest competing logit.} If \(\Delta_\theta(x,x')<m_\theta(x)/2\), the prediction is preserved under the perturbation.} 
\rev{This gives a sufficient condition for classification stability, with the derivation provided in Supplementary Materials S1-\textit{B}.}

\rev{Under this view, both BSNs and DWC improve robustness by reducing different sources of output variation \(\Delta_\theta(x,x')\). BSNs reduce perturbation-induced activation-state changes, while DWC constrains the weights that propagate such changes across layers.}


\subsection{Preliminaries of Spiking Neuron Models}
In general, the forward dynamics of an SNN neuron can be abstracted as a three-stage process comprising membrane potential update (charge), spike generation (fire), and state reset (reset), which can be formulated as follows:
\begin{align}
    V[t] &= \mathcal{U}(V[t{-}1], I[t]), \label{eq:charge} \\ 
    S[t] &= \Theta(V[t] - V_{\text{th}}), \label{eq:fire} \\
    V[t] &= 
    \begin{cases} 
        V[t] (1 - S[t]) + V_{\text{reset}} S[t], & \text{hard reset}, \\  
        V[t] - V_{\text{th}} S[t], & \text{soft reset},
    \end{cases} \label{eq:reset}
\end{align}
where \( V[t] \in \mathbb{R} \) is the membrane potential at time step \( t \), \( I[t] \) is the synaptic input current, and \( S[t] \in \{0, 1\} \) denotes the binary spike output. The function \( \mathcal{U}(\cdot) \) defines neuron dynamics, and \(\Theta(\cdot)\) emits a spike when \(V[t]\) exceeds the threshold \(V_{\text{th}}\). Once a spike occurs, the membrane potential is reset according to Eq.~\eqref{eq:reset}, either to the reset potential \(V_{\text{reset}}\) or by subtracting the threshold \(V_{\text{th}}\).

\rev{A variety of models instantiate the neuron dynamic \( \mathcal{U}(\cdot) \), such as LIF~\cite{lif}, PLIF~\cite{plif}, and GLIF~\cite{glif}. Among them, the LIF model updates its membrane potential recursively with temporal decay:}
\begin{align}
\mathcal{U}(V[t{-}1], I[t]) = \left(1 - \frac{1}{\tau_m}\right)V[t-1] + \frac{1}{\tau_m} I[t], \label{eq:lif_recursive}
\end{align}
where \( \tau_m \) is the membrane time constant. \rev{This recursive form depends on \( V[t{-}1] \) and therefore limits parallel computation across time steps.}
To address this limitation, PSN~\cite{psn} reformulates the membrane potential with a closed-form update, which unrolls the recursion into a temporal convolution over all past inputs:
\begin{align}
V[t] = \frac{1}{\tau_m} \sum_{i=0}^{t} \left(1 - \frac{1}{\tau_m}\right)^{t - i} I[i].\label{eq:lif_closed}
\end{align}
\rev{For an input sequence with \(T\) time steps and \(N\) neurons, PSN defines the membrane potential as a weighted summation over all historical input currents}:
\begin{align}
V[t] = \sum_{i=0}^{T-1} W_{t,i} \cdot I[i], \label{eq:psn_sum}
\end{align}
where the temporal kernel \( W_{t,i} = \frac{1}{\tau_m} \left(1 - \frac{1}{\tau_m} \right)^{t - i} \) encodes exponentially decaying influence over past inputs. By incorporating the firing process in Eq.~\eqref{eq:fire} and omitting the reset in Eq.~\eqref{eq:reset}, PSN expresses neuron dynamics in a matrix formulation:
\begin{equation}
\left\{
\begin{aligned}
    \mathbf{V} &= \mathbf{W} \mathbf{I}, \quad \mathbf{W} \in \mathbb{R}^{T \times T},\ \mathbf{I} \in \mathbb{R}^{T \times N}, \\
    \mathbf{S} &= \Theta(\mathbf{V} - \mathbf{B}), \quad \mathbf{B} \in \mathbb{R}^{T},\ \mathbf{S} \in \{0, 1\}^{T \times N},
\end{aligned}
\right.
\label{eq:psn}
\end{equation}
where \( \mathbf{I} \) denotes the input current sequence, \( \mathbf{V} \) the membrane potential, \( \mathbf{B} \) a learnable threshold vector, and \( \mathbf{S} \) the spike output. This feedforward formulation removes recurrent dependencies and enables parallel evaluation across time steps, thereby improving training and inference efficiency.

The Memory-based Parallel Spiking Neuron (MPSN)~\cite{mpsn} extends PSN by introducing a memory-based modulation. Instead of directly using \( \mathbf{I} \), the input current is dynamically modulated by a learned memory unit \( \mathbf{M} \in \mathbb{R}^{T \times N} \):
\begin{equation}
\left\{
\begin{aligned}
    \mathbf{V} &= \mathbf{W} \, g(\mathbf{M}, \mathbf{I}), \quad \mathbf{W} \in \mathbb{R}^{T \times T},\ \mathbf{I}, \mathbf{M} \in \mathbb{R}^{T \times N}, \\
    \mathbf{S} &= \Theta(\mathbf{V} - \mathbf{B}), \quad \mathbf{B} \in \mathbb{R}^{T},\ \mathbf{S} \in \{0, 1\}^{T \times N},
\end{aligned}
\right.
\label{eq:pmsn6}
\end{equation}
where the modulation function is defined as \( g(\mathbf{M}, \mathbf{I}) = \mathbf{I} + \mu \mathbf{M} \), and \( \mu \) is a scaling factor, with a default value of 0.1. This memory mechanism enables the model to capture statistically stable features from clean samples, thereby improving robustness to noisy or corrupted inputs.
\rev{In this work, we adopt MPSN as the base neuron model and extend its binary spike to burst spike.}

\subsection{Burst-enhanced Spiking Neurons}
Conventional SNN neurons typically produce binary outputs, which are insufficient to encode fine-grained input intensities or to distinguish between similar stimuli, potentially impairing accuracy and robustness. In contrast, biological neurons frequently exhibit burst spiking, which refers to short sequences of spikes emitted in rapid succession, as illustrated in Fig.~\ref{fig:fig1} (b). This spiking pattern is widely observed across brain regions~\cite{krahe2004burst}. It is believed to enhance the reliability of information transmission, thereby improving robustness to input perturbations~\cite{lisman1997bursts,izhikevich2000neural}.

\subsubsection{Biological modeling and burst definition} 
Biologically, burst spiking is a firing pattern in which a neuron emits brief clusters of high-frequency action potentials, producing multiple spikes within a short time window. 
Let \(s(t)\) denote the spike train in continuous time, and let \(\Delta t_b>0\) the burst window. 
A burst event starting at time \(t\) can be formulated as:
\begin{align}
\mathrm{Burst}(t) = \{\, \tau \mid s(\tau)=1,\ t \le \tau < t+\Delta t_b \,\}.
\label{eq:burst_set_ct}
\end{align}
\rev{In general, burst activity can convey information through multiple factors, such as burst onset time, within-burst spike timing, and the number of spikes within the burst window.}

\subsubsection{From biological mechanisms to computational abstractions}
SNNs for computer vision commonly adopt a coding paradigm in which neuronal activation is summarized by firing rates over time. 
Motivated by voltage-based spike generation models~\cite{gerstner2002spiking} and multi-bit spiking formulations~\cite{luo2024integer,yao2025scaling}, we use spike-count coding to represent burst spiking. 
The spike count $N_{\text{burst}}[t]$ can be approximated by the membrane potential surplus above the firing threshold. We therefore define a deterministic mapping from the membrane potential \(V[t]\) to $N_{\text{burst}}[t]$:
\begin{align}
    N_{\text{burst}}[t]
    =
    \mathcal{B}\!\left(V[t] - V_{\text{th}}\right),
    \label{eq:burst_count}
\end{align}
where \(\mathcal{B}:\mathbb{R}\to\{0,1,\dots,\theta_{\text{burst}}\}\) is a quantized linear mapping of the excess potential:
\begin{align}
    \mathcal{B}(x) \;=\; \min\!\left( \theta_{\text{burst}},\;\left\lfloor \max(x,0) + \frac{1}{2} \right\rfloor \right). \label{eq:burst_spiking_neuron}
\end{align}
Here \(x\) denotes the membrane surplus \(V[t]-V_{\text{th}}\) measured in normalized threshold-equivalent units, and \(\theta_{\text{burst}}\) specifies the maximum number of spikes allowed in each burst to ensure stability. Unless otherwise stated, \(\theta_{\text{burst}}\) is set to 4 in all experiments. \rev{Under this normalization, each additional unit of surplus results in an additional spike emission within the burst window. This form provides a bridge from the biological burst count to a discrete response that can be computed in SNNs.}

\subsubsection{Neuron model}
We propose BSNs \rev{(Fig.~\ref{fig:fig1} (c))} based on the MPSN model in Eq.~\eqref{eq:pmsn6} by incorporating the burst spiking in Eq.~\eqref{eq:burst_spiking_neuron} into it. While the burst-spiking mechanism can be integrated into other models, we adopt MPSN for its strong performance. The BSN is defined as follows:
\begin{align}
\left\{
\begin{aligned}
    \mathbf{V} &= \mathbf{W} \, g(\mathbf{M}, \mathbf{I}), \quad \mathbf{W} \in \mathbb{R}^{T \times T},\ \mathbf{I}, \mathbf{M} \in \mathbb{R}^{T \times N}, \\
    \mathbf{S} &= \mathcal{B}(\mathbf{V} - \mathbf{B}), \quad \mathbf{B} \in \mathbb{R}^{T},\ \mathbf{S} \in \{0,1,\dots,\theta_{\text{burst}}\}^{T \times N}.
\end{aligned}
\right.
\label{eq:bsn}
\end{align}
When BSNs are trained end-to-end, the \rev{burst mapping} \(\mathcal{B}(\cdot)\) is non-differentiable at integer boundaries and piecewise constant elsewhere. We therefore use a smooth arctangent-shaped surrogate in the backward pass:
\begin{equation}
\frac{\partial \mathcal{B}}{\partial x} \approx \sigma(x)
= \frac{\alpha}{1+(\alpha x)^2}\,\boldsymbol{1}_{\big(0 < x < \theta_{\text{burst}}\big)},
\end{equation}
where \(\alpha=4\), and \(\boldsymbol{1}_{(\cdot)}\) denotes the indicator function. Compared with standard binary spiking neurons, BSNs provide a richer output space while preserving the parallel computation pattern of the parallel spiking neuron family.

\subsubsection{\rev{Robustness Interpretation of Burst Spiking}}
\rev{Eq.~\eqref{eq:delta_output} measures the output variation caused by input perturbations. At the neuron level, such variation is partly induced by perturbation-driven changes in the membrane surplus. BSNs reduce this effect through the burst response in Eq.~\eqref{eq:burst_spiking_neuron}. Let $u$ denote the normalized membrane surplus, and let $\mathcal{Q}=\{k+\frac{1}{2} \mid k=0,\dots,\theta_{\text{burst}}-1\}$ denote the set of burst-response boundaries induced by $\mathcal{B}$. We define the distance from $u$ to the nearest response boundary as $d_{\mathcal{Q}}(u)=\min_{q\in\mathcal{Q}} |u-q|$. If $|\delta|<d_{\mathcal{Q}}(u)$, then $u+\delta$ stays in the same response interval and $\mathcal{B}(u+\delta)=\mathcal{B}(u)$. If the perturbation crosses response boundaries, the burst-count change is bounded by the number of unit-spaced response intervals crossed:
\begin{equation}
\bigl|\mathcal{B}(u+\delta)-\mathcal{B}(u)\bigr|
\le
\left\lceil|\delta|\right\rceil .
\label{eq:burst_stability}
\end{equation}
}
\rev{These properties allow BSNs to suppress perturbations that remain within the same response interval and limit the activation change when response boundaries are crossed. Under the output-variation view in Eq.~\eqref{eq:delta_output}, this reduces the activation perturbation propagated to later layers. Theorem 1 in Supplementary Materials S1-\textit{C} formalizes this activation-side effect by showing how the burst-count change in Eq.~\eqref{eq:burst_stability} determines the perturbation signal received by the next layer. BSNs therefore act on the activation side.}

\subsection{Dynamic Weight Constraint}

Robustness to input perturbations can be intuitively understood from the perspective of network connectivity.  
Consider a simple linear mapping \(y = Wx + b\).  
Smaller values of connection weights \(W\) reduce the sensitivity of the output \(y\) to perturbations in the input \(x\). 
However, excessively reducing the weights often degrades model performance, indicating the need for a well-designed mechanism to regulate weight magnitudes.
Such regulation naturally occurs in biological neural systems through synaptic plasticity, where synaptic strengths are adjusted under various biological constraints.

\subsubsection{Competitive and constrained synaptic plasticity}

In biological neural systems, synaptic connections evolve under limited metabolic resources, structural capacity, and homeostatic balance. These constraints help maintain efficient information processing and prevent unstable growth of synaptic strength. From this perspective, we focus on two mechanisms, \textit{synaptic competition} and \textit{synaptic constraint}.

Synaptic competition refers to the process in which synapses compete for survival and efficacy based on their relative activity levels~\cite{changeux1976selective, purves1980elimination}. 
Active synapses that contribute effectively to postsynaptic responses are strengthened and preserved, whereas less active or redundant synapses are gradually weakened and eliminated~\cite{ooyen2001competition}. 
\rev{This process can be modeled as:}
\begin{equation}
\frac{d w_i}{dt} = p_i \cdot h(w_i) - \lambda \sum_{j \ne i} p_j \cdot u(w_j), \label{eq26}
\end{equation}
where \(w_i\) denotes the weight of synapse \(i\), \(p_i\) represents its activity level, and \(h(\cdot)\) and \(u(\cdot)\) govern potentiation and competitive suppression, respectively. 
This formulation captures how the strengthening of one synapse can inhibit others, leading to sparse and selective connectivity.

Synaptic constraint reflects the fact that neurons cannot increase all synaptic strengths without limit.
\rev{
Let $w_i\ge0$ denote the strength of synapse $i$.
\rev{Assuming a positive total synaptic strength $\sum_j w_j>0$,}
this constraint can be described by normalization-like dynamics:
\begin{equation}
\frac{d w_i}{dt}
=
\eta
\left(
\rho_i
-
\frac{w_i}{\sum_j w_j}
\right),
\qquad
\rho_i\ge0,\quad \sum_i\rho_i=1,
\label{eq28}
\end{equation}
where $\rho_i$ is the normalized activity-dependent target share.
The term $w_i/\sum_j w_j$ denotes the current share of synapse $i$ in the total synaptic strength, \rev{i.e., the relative fraction of the available synaptic-strength budget assigned to this synapse rather than its absolute magnitude}.
\rev{Thus, the dynamics compare the target share $\rho_i$ with the current normalized share $w_i/\sum_j w_j$. A synapse is strengthened when its current share is below the target, and weakened when it exceeds the target.}
This prevents \rev{unbounded growth of all synapses and limits} disproportionate dominance among active synapses, thereby helping maintain stable connectivity~\cite{miller1994role}.
\rev{Since $\sum_i\rho_i=1$ and $\sum_i w_i/\sum_j w_j=1$, the total synaptic strength is conserved under Eq.~\eqref{eq28}.}
At equilibrium, we have:
\begin{equation}
w_i^\ast = \rho_i Q,
\label{eq30}
\end{equation}
where $Q=\sum_jw_j^\ast$ is the total synaptic-strength budget \rev{determined by the conserved total strength}.
Eq.~\eqref{eq30} indicates that, under a constrained total budget, the equilibrium synaptic strength is proportional to activity.
}


\rev{These two principles motivate the design of DWC.
The competition principle motivates a penalty on inactive or weakly active connections with large weights.
The constraint principle motivates an activity-weighted penalty on active connections.}

\subsubsection{From biological mechanisms to regularization formulation}  

Directly modeling these biological mechanisms is difficult in gradient-based SNN training.
\rev{We therefore translate the core principles into a weight-regularization framework, termed DWC. It contains two complementary terms, as shown in Fig.~\ref{fig:fig1}(d). The synaptic competition suppresses inactive connections with excessively large weights. The synaptic constraint regularizes active connections according to their activation strength. Together, they reduce redundant or unstable connections and lower weight-side sensitivity to perturbations.}

To formulate the synaptic competition model in Eq.~\eqref{eq26}, we define \( w_i=h(w_i)  \) to represent Hebbian-like linear potentiation and \( |w_j|=u(w_j) \) to capture the suppressive influence of stronger neighboring synapses:
\begin{equation}
\frac{d w_i}{dt} = p_i w_i - \lambda \sum_{j \ne i} p_j |w_j|.
\end{equation}
For inactive connections (\(p_i = 0\)), the potentiation term vanishes, and the dynamics reduce to a purely inhibitory drift proportional to the total active mass in the neighborhood:
\begin{equation}
\label{eq:inactive_dynamics}
\frac{d w_i}{dt} = -\,\lambda \sum_{j \ne i} p_j |w_j|.
\end{equation}

To capture this inhibitory effect, we introduce a static convex penalty that enforces a layerwise reference level determined by the active channels.  
For an input \(x\), let \(\mathcal{I}^{(l)}_x\) denote the set of {active} channels at layer \(l\), and \(\overline{\mathcal{I}^{(l)}_x}\) its complement (the {inactive} channels).
For layer \(l\), define the maximum weight norm among active channels as:
\begin{equation}
\label{eq:M_def}
M^{(l)}(x) =
\begin{cases}
\max\limits_{k \in \mathcal{I}^{(l)}_x} \| W^{(l)}_k \|_1, & \mathcal{I}^{(l)}_x \neq \varnothing,\\[2pt]
0, & \mathcal{I}^{(l)}_x = \varnothing.
\end{cases}
\end{equation}
Based on this active reference, we define the following competitive penalty on inactive connections. \rev{This penalty mainly affects inactive channels with large weights}:
\begin{equation}
\label{eq:Pcomp_layer}
\mathcal{P}^{(l)}_{\text{comp}}(x)
= \sum_{c \in \overline{\mathcal{I}^{(l)}_x}}
\max\!\bigl( \| W^{(l)}_c \|_1,\; M^{(l)}(x) \bigr).
\end{equation}
Aggregating over layers gives:
\begin{equation}
\label{eq:Pcomp_total}
\mathcal{P}_{\text{comp}}(x)
= \sum_{l=1}^{L} \mathcal{P}^{(l)}_{\text{comp}}(x).
\end{equation}

\rev{
\rev{Eq.~\eqref{eq30} indicates that, under a constrained total synaptic budget \(Q=\sum_j w_j\), the equilibrium strength of each synapse is proportional to its activity share.
Here \(w_i\) denotes a non-negative biological synaptic strength. In deep neural networks, however, trainable weights are signed, and the corresponding synaptic strength is more naturally represented by the weight magnitude.
Thus, a direct constraint can be written as
\(\sum_i \left||w_i|-\rho_i\sum_j |w_j|\right|\).}
In practice, this objective couples all synapses through the global budget
term \(\sum_j |w_j|\), which is inconvenient for stochastic gradient training.
We therefore use a tractable activity-weighted surrogate $\sum_i a_i |w_i|$. Here \(a_i\) denotes the activation-dependent coefficient of synapse \(i\), and it can be viewed as the computational counterpart of \(\rho_i\). This term is not an algebraically exact reformulation of Eq.~\eqref{eq30}, but a separable convex proxy that preserves its key effect: weights associated with stronger activity are more explicitly constrained during training.
The derivation of this regularization form is detailed in Supplementary Materials S1-\textit{D}}.


Regrouping {$\sum_i a_i |w_i|$} by presynaptic unit gives:
\begin{equation}
\mathcal{P}_{\text{cons}}(x) = \sum_{c \in \mathcal{I}_x} a_c\, \|W_c\|_1,
\label{eq:pcons}
\end{equation}
where \(c\) denotes a presynaptic unit, \(W_c\) denotes its associated outgoing weights, and \(a_c\) is the corresponding activation coefficient. 
For binary spiking neurons, \(a_c=1\) for active units \(c\in\mathcal{I}_x\), and the penalty reduces to a standard \(\ell_1\) constraint on active connections. 
For BSNs, \(a_c\) reflects the burst activation strength, enabling the weight constraint to adapt to the activation state. Thus, Eq.~\ref{eq:pcons} regularizes the weights involved in the current computation in an activation-dependent manner, discouraging the model from relying on excessively large active weights.

Combining \(\mathcal{P}_{\text{comp}}(x)\) and \(\mathcal{P}_{\text{cons}}(x)\) yields the overall DWC loss:
\begin{equation}
\mathcal{L}_{\text{dwc}}(x)
= \sum_{l=1}^{L} \sum_{c \in \overline{\mathcal{I}^{(l)}_x}}
\max\!\bigl( \| W^{(l)}_c \|_1, \, M^{(l)}(x) \bigr) + \sum_{c \in \mathcal{I}_x} a_c \|W_c\|_1.
\label{eq:eq34}
\end{equation}

\rev{
DWC imposes activation-conditioned penalties on network weights.
For active channels, the BSN response gives \(a_c\ge1\), while inactive channels are constrained by the competitive term.
Therefore, DWC preserves the magnitude-control effect of standard \(\ell_1\) regularization.
More importantly, it adapts this penalty to the input-dependent activation pattern, so that active or potentially sensitive connections are more directly constrained.
This helps reduce weight-side amplification of perturbations.
}

Finally, the network is trained using a composite loss function:
\begin{equation}
\mathcal{L}_{\text{total}} = \mathcal{L}_{\text{cls}} + \mathcal{L}_{\text{dwc}},
\end{equation}
where $\mathcal{L}_{\text{cls}}$ is the classification loss.

\subsubsection{\color{black}{Robustness Interpretation of DWC}}
\label{method:D.3}

\rev{
Following the output-stability definition in Eq.~\eqref{eq:delta_output}, robustness depends on how much the network output \(f_\theta(x)\) changes under a perturbed input \(x'\).
BSNs reduce such changes by stabilizing neural activations.
DWC further limits how the remaining activation changes are propagated through network weights.
For a linear transformation in layer \(l\), this propagation satisfies:}
{\color{black}
\begin{equation}
\begin{aligned}
&\left\|W^{(l)}
\left(\mathbf{S} ^{(l-1)}(x')-\mathbf{S} ^{(l-1)}(x)\right)
\right\|_\infty  \\
&\qquad\le
\left\|W^{(l)}\right\|_\infty
\left\|\mathbf{S} ^{(l-1)}(x')-\mathbf{S} ^{(l-1)}(x)\right\|_\infty .
\end{aligned}
\label{eq:weight_amplification_bound}
\end{equation}
}
\rev{
Here, \(\mathbf{S} ^{(l)}(\cdot)\) denotes the spike-count activation of layer \(l\), as defined in the BSN forward model in Eq.~\eqref{eq:bsn}, and \(W^{(l)}\) denotes the weight matrix or convolutional kernel of layer \(l\), as used in the DWC formulation in Eq.~\eqref{eq:eq34}. \rev{Eq.~\eqref{eq:weight_amplification_bound}} shows that smaller weight norms reduce the amplification of perturbation-induced activation changes.
DWC regularizes these weights in an activation-dependent manner.
The active term constrains weights involved in the current computation, while the inactive term suppresses large dormant weights that may become influential when perturbations change activation states.
Thus, DWC reduces the weight-side amplification that contributes to the output variation \(\Delta_\theta(x,x')\) in Eq.~\eqref{eq:delta_output}.
}

\rev{
In summary, BSNs stabilize neuronal activations under perturbations, whereas DWC reduces weight-side amplification by constraining weight magnitudes.
Together, they reduce the output variation of BuSNNs and contribute to improved robustness.
}

\begin{table}
    \centering
    \caption{Training hyper-parameters for different datasets.}
    \begin{tabularx}{0.48\textwidth}{l c c c c c}
        \toprule
        Dataset & Optimizer & Batch & Epoch & LR & Loss \\
        \midrule
        CIFAR-10 & SGD & 128 & 1024 & 0.1 & CE \\
        CIFAR-100 & SGD & 128 & 1024 & 0.1 & CE \\
        ImageNet & SGD & 64 & 320 & 0.001 & TET \\
        CIFAR10-DVS & SGD & 32 & 200 & 0.1 & TET \\
        \bottomrule
    \end{tabularx}
    \label{tab:hyperparameters}
\end{table}

\section{Experiments}

\subsection{Experimental Settings}

We evaluate our proposed method across multiple datasets to assess both classification accuracy and robustness. All experiments are conducted using the SpikingJelly framework~\cite{fang2023spikingjelly}. We use CIFAR-10~\cite{cifar10}, CIFAR-100~\cite{cifar10}, and ImageNet~\cite{imagenet} as the primary benchmarks to evaluate the general performance of the models. To examine adversarial robustness, we follow established protocols and evaluate on adversarially perturbed versions of CIFAR-10 and CIFAR-100. Furthermore, to assess corruption robustness, we adopt the common benchmarks CIFAR10-C~\cite{hendrycks2019benchmarking} and IMAGENET-C~\cite{hendrycks2019benchmarking}, which apply a wide range of corruptions such as noise, blur, and weather distortions to the original datasets. To ensure a thorough comparison with prior methods, we conduct comprehensive experiments using a range of network architectures, including ResNet~\cite{ResNet18}, SEW ResNet~\cite{sew}, MS ResNet~\cite{hu2024advancing}, VGG~\cite{vgg}, and the Modified PLIF-Net~\cite{fang2021incorporating}. 

\begin{table}[t]
  \caption{Comparison of the proposed BuSNN and other methods on the CIFAR-10 dataset. \rev{\underline{Underlined} values indicate the best performance.}}
  \label{tab:cl-cifar}
  \centering
  \begin{tabular}{l p{2.7cm} c c}
    \toprule
    Method & Network & Time-steps & Accuracy (\%) \\
    \midrule
    ANN     & Modified PLIF-Net & -  & 95.84  \\
    \midrule
    LIF     & Modified PLIF-Net & 6   & 95.04  \\
    TET \cite{tet}     & ResNet-19 & 6   & 94.50  \\
    TDBN \cite{tdbn}   & ResNet-19       & 6  & 93.16   \\
    TEBN  \cite{tebn}  & ResNet-19   & 6   & 94.71 \\
    PLIF \cite{plif}   & PLIF Net     & 8  & 93.50 \\
    KLIF \cite{klif}   & Modified PLIF-Net    & 10  & 92.52 \\
    GLIF \cite{glif}   & ResNet-19     & 6  & 95.03 \\
    NLIF \cite{nsnn}   & ResNet-18      & 4  & 93.73 \\
    SLTT \cite{sltt}   & Modified PLIF-Net    & 4  & 95.08 \\ 
    CLIF \cite{clif}   & ResNet-18    & 4  & 96.01 \\ 
    LTF-ALSF \cite{fu2025adaptation} & Modified PLIF-Net    & 4  & 94.48 \\ 
    PSN \cite{psn}     & Modified PLIF-Net    & 4  & 95.32 \\
    MPSN \cite{mpsn}   & Modified PLIF-Net    & 4  & 96.18 \\
    BuSNN (ours)        & Modified PLIF-Net   & 4  & \underline{\textbf{96.25}} \\
    \bottomrule
  \end{tabular}
\end{table}

\begin{table}[t]
  \caption{Comparison of the proposed BuSNN and other methods on the ImageNet dataset.  Accuracy is top-1 (\%). For ANN rows, `a / b' denotes Full-precision / INT4 activation-quantized results. \rev{\underline{Underlined} results correspond to the optimal performance among all compared SNN models.}}
  \label{tab:cl-imagenet}
  \centering
  \begin{tabular}{l p{2.5cm} c c}
    \toprule
    Method & Spiking Network & Time-steps & Accuracy (\%) \\
    \midrule
    TET \cite{tet}     & SEW ResNet-34 & 4   & 68.00  \\
    TDBN \cite{tdbn}   & ResNet-34 & 6   & 67.05   \\
    TEBN \cite{tebn}   & SEW ResNet-34 & 4   & 68.28 \\
    GLIF \cite{glif}   & ResNet-34     & 4  & 67.52\\
    NLIF \cite{nsnn}   & SEW ResNet-18 & 4  & 63.32 \\
    LTF-ALSF \cite{fu2025adaptation} & ResNet-18 & 4 & 63.67 \\
   \cmidrule(r){2-4}
     \multirow{2}[4]{*}{ANN \cite{ResNet18}} & ResNet-18 & -  & 69.76 / 66.44 \\
                       & ResNet-34 & -  & 73.31 / 69.73 \\
                       & ResNet-50 & -  & 76.13 / 71.05 \\ 
   \cmidrule(r){2-4}
     \multirow{2}[4]{*}{SEW \cite{sew}} & SEW ResNet-18 & 4  & 63.18 \\
                       & SEW ResNet-34 & 4  & 67.04 \\
                       & SEW ResNet-50 & 4  & 67.78 \\ 
    \cmidrule(r){2-4}
     \multirow{2}[4]{*}{MS \cite{hu2024advancing}} & MS ResNet-18 & 6  & 63.10 \\
                       & MS ResNet-34 & 6  & 69.42 \\
                       & MS ResNet-50 & 6  & 71.05 \\ 
   \cmidrule(r){2-4}
   \multirow{2}[4]{*}{PSN \cite{psn}}
                        & SEW ResNet-18 & 4  & 67.63 \\
                       & SEW ResNet-34 & 4  & 70.54 \\  
                       & SEW ResNet-50 & 4  & 70.75 \\
   \cmidrule(r){2-4}
   \multirow{2}[4]{*}{MPSN \cite{mpsn}}
                       & SEW ResNet-18 & 4  & 67.92 \\
                       & SEW ResNet-34 & 4  & 71.30 \\
                       & SEW ResNet-50 & 4  & 71.56 \\
   \cmidrule(r){2-4}
   \multirow{2}[4]{*}{BuSNN (ours) }
                        & SEW ResNet-18 & 4  & \underline{\textbf{69.92}} \\
                       & SEW ResNet-34 & 4  & \textbf{72.40} \\
                       & SEW ResNet-50 & 4  & \textbf{72.70} \\
    \cmidrule(r){2-4}
   \multirow{2}[4]{*}{BuSNN (ours) }
                        & MS ResNet-18 & 4  & \textbf{69.64} \\
                       & MS ResNet-34 & 4  & \underline{\textbf{72.60}} \\
                       & MS ResNet-50 & 4  & \underline{\textbf{73.70}} \\
    \bottomrule
  \end{tabular}
\end{table}

The hyperparameters used in our experiments are listed in Table~\ref{tab:hyperparameters}, where CE and TET refer to the cross-entropy loss and temporal efficient training~\cite{tet}, respectively, used as the classification loss. We optimize the models using SGD with a momentum of 0.9 and apply a cosine annealing schedule for the learning rate.

To ensure a fair comparison, we adopt the same data preprocessing methods as used in PSN~\cite{psn} and MPSN~\cite{mpsn}.
\rev{All experiments are implemented with fixed random seeds to facilitate reproducibility. For CIFAR-10 and CIFAR-100, we use random horizontal flipping and random erasing. For each mini-batch, we randomly apply either Mixup~\cite{zhang2017mixup} with \(\alpha=0.2\) or CutMix~\cite{yun2019cutmix} with \(\alpha=1.0\), where \(\alpha\) controls the sampled mixing ratio. For ImageNet, we follow the conventional protocol with random resized cropping and horizontal flipping. We initialize our model with MPSN pre-trained weights to provide a better starting point for training.} Similar strategies that leverage stronger initialization have been widely adopted to improve convergence and generalization~\cite{rathi2019enabling,rathi2021diet,mpsn}.
\rev{As shown in Table~\ref{tab:cl-cifar} and \ref{tab:cl-imagenet}, for SNN baselines from prior literature, we follow the time-step settings reported in the original papers. Unless otherwise specified, all experiments reported in Tables~\ref{tab:Adversarial_Attack}--~\ref{tab:ablation_burst} are conducted with the number of time steps set to \(T=4\).}

\subsection{Network Architectures}
We adopt the networks with the same settings used in MPSN~\cite{mpsn}, including Modified PLIF-Net~\cite{fang2021incorporating}, VGG~\cite{vgg}, and ResNet~\cite{ResNet18}. As illustrated in Fig.~\ref{fig:fig1} (a), these networks share a common pattern in which image features are extracted by a stack of convolutional (Conv) blocks. Each block comprises convolutional layers followed by a nonlinear activation. In SNNs, the nonlinearity is implemented with spiking neurons, whereas in ANNs it is implemented with ReLU. The classifier consists of a stack of fully connected (FC) layers. Detailed configurations are provided in Supplementary Materials S2. In our BuSNN variants, we replace the original spiking neurons in these backbones with BSNs and apply DWC to the convolutional and fully connected layers, illustrating that the proposed modules can be directly used in existing SNN architectures.

\begin{table*}[t]
    \caption{\rev{Comparison of recognition accuracy (\%) of different methods under FGSM, PGD7 and PGD10 adversarial attacks. FGSM uses a single update step, while PGD7 and PGD10 adopt 7 and 10 iterations respectively. BuSNN- denotes the BuSNN without using DWC. All values of our proposed methods are \textbf{bolded}, and the best performance in each column within each group is \underline{underlined}.}}
    \centering
    \begin{tabularx}{\textwidth}{@{}ll>{\hsize=2\hsize}XX|XX|XX|XXX@{}}
    \toprule
         Dataset&  Method&  Spiking Network & Clean & FGSM & \textcolor{black}{Score} &PGD7 & \textcolor{black}{Score} &PGD10 & \textcolor{black}{Score} \\
         \midrule
         \multirow{14}{*}{CIFAR-10} 
         & ANN & Modified PLIF-Net & 95.84 & 61.05 & \textcolor{black}{0.6370} & \underline{15.25} & \textcolor{black}{\underline{0.1591}} & \underline{13.71} & \textcolor{black}{\underline{0.1431}} \\
         \cmidrule(r){2-10}
         & PSN \cite{psn} & Modified PLIF-Net & 95.32 & 61.29 & \textcolor{black}{0.6430} & 9.61 & \textcolor{black}{0.1008} & 8.04 & \textcolor{black}{0.0843}\\
         & MPSN \cite{mpsn}& Modified PLIF-Net & 96.18 & 63.11 & \textcolor{black}{0.6562} & 12.81 & \textcolor{black}{0.1332} & 11.56 &\textcolor{black}{0.1202} \\
         & BuSNN- (ours) & Modified PLIF-Net & \textbf{96.22} & \textbf{\underline{68.77}} & \textcolor{black}{\underline{\textbf{0.7147}}} & \textbf{14.73} & \textcolor{black}{\textbf{0.1531}} & \textbf{13.03} & \textcolor{black}{\textbf{0.1354}}\\
         & BuSNN (ours) & Modified PLIF-Net & \textbf{\underline{96.25}} & \textbf{67.89} & \textcolor{black}{\textbf{0.7054}} & \textbf{14.62} & \textcolor{black}{\textbf{0.1519}} & \textbf{13.23} & \textcolor{black}{\textbf{0.1375}}\\
         \cmidrule(r){2-10}
         & ANN & VGG11 & 95.92 & 71.97 & \textcolor{black}{0.7503} & \underline{30.73} & \textcolor{black}{\underline{0.3204}} & \underline{28.72} & \textcolor{black}{\underline{0.2994}} \\
         \cmidrule(r){2-10}
         & DLIF \cite{ding2024robust} & VGG11 & 92.39 & 15.24 & \textcolor{black}{0.1650} & 0.23 & \textcolor{black}{0.0025} & 0.09 & \textcolor{black}{0.0010}\\
         & PSN \cite{psn} & VGG11 & 95.21 & 63.30 & \textcolor{black}{0.6648} & 24.14 & \textcolor{black}{0.2535} & 22.27 & \textcolor{black}{0.2339} \\
         & MPSN \cite{mpsn} & VGG11 & 95.17 & 65.02 & \textcolor{black}{0.6832} & 24.27 & \textcolor{black}{0.2550} & 22.19 & \textcolor{black}{0.2332}\\
         & BuSNN- (ours) & VGG11 & \textbf{96.01} & \textbf{71.64} & \textcolor{black}{\textbf{0.7462}} & \textbf{29.24} & \textcolor{black}{\textbf{0.3046}} & \textbf{26.83} & \textcolor{black}{\textbf{0.2795}} \\
         & BuSNN (ours) & VGG11 & \textbf{\underline{96.10}}  & \textbf{\underline{72.15}} & \textcolor{black}{\underline{\textbf{0.7508}}} & \textbf{29.03} & \textcolor{black}{\textbf{0.3021}} & \textbf{27.01} & \textcolor{black}{\textbf{0.2811}} \\
         \midrule
         \multirow{8}[4]{*}{CIFAR-100} 
         & ANN & VGG11 & 70.44 & 23.52 & \textcolor{black}{0.3339} & 8.08 & \textcolor{black}{0.1147} & 7.67 & \textcolor{black}{0.1089}\\
         \cmidrule(r){2-10}
         & DLIF \cite{ding2024robust} & VGG11 & 70.51 & 8.72 & \textcolor{black}{0.1237} & 0.77 & \textcolor{black}{0.0109} & 0.55 & \textcolor{black}{0.0078}\\
         & StoG \cite{ding2024enhancing} & VGG11 & 70.44 & 8.27 & \textcolor{black}{0.1174} & 0.49 &\textcolor{black}{0.0070} & - & -\\
         & HIRE-SNN \cite{kundu2021hire} & VGG11 & 65.6 & 16.4 & \textcolor{black}{0.2500} & 2.9 & \textcolor{black}{0.0442} & - & - \\
         & SNN-BP \cite{sharmin2020inherent} & VGG11 & 64.4 & 15.5 & \textcolor{black}{0.2407} & 6.3 & \textcolor{black}{0.0978} & - & -\\
         & PSN \cite{psn} & VGG11 & 65.03 & 18.35 & \textcolor{black}{0.2822} & 4.79 & \textcolor{black}{0.0737} & 4.17 & \textcolor{black}{0.0641}\\
         & BuSNN- (ours) & VGG11 & \textbf{71.01} & \textbf{20.84} & \textcolor{black}{\textbf{0.2935}} & \textbf{4.93} & \textcolor{black}{\textbf{0.0694}} & \textbf{4.61} & \textcolor{black}{\textbf{0.0649}}\\
         & BuSNN (ours) & VGG11 & \textbf{\underline{71.23}} & \textbf{\underline{29.90}} & \textcolor{black}{\underline{\textbf{0.4198}}} & \textbf{\underline{12.97}} & \textcolor{black}{\underline{\textbf{0.1821}}} & \textbf{\underline{12.43}} & \textcolor{black}{\underline{\textbf{0.1745}}}\\
         \bottomrule
    \end{tabularx}
    \label{tab:Adversarial_Attack}
\end{table*}

\subsection{Results on Clean Images}

To evaluate the proposed model, we conduct experiments on the clean image classification benchmarks CIFAR-10~\cite{cifar10} and ImageNet~\cite{imagenet}. On both datasets, the proposed BuSNN achieves superior performance among existing SNNs, as listed in Table~\ref{tab:cl-cifar} and Table~\ref{tab:cl-imagenet}. 

Compared with ANN baselines, BuSNN achieves stronger performance on CIFAR-10. On ImageNet, the SEW ResNets \cite{sew} and MS ResNet~\cite{hu2024advancing} architectures serve as the spiking counterpart of the standard ANN ResNets \cite{ResNet18} and have been widely adopted in the design of residual SNNs. BuSNN with the SEW ResNet-18 backbone also surpasses its ANN counterpart (69.92\% vs.\ 69.76\%). Although a gap remains on ResNet-34 and ResNet-50, our approach substantially narrows the difference relative to previous SNN methods, demonstrating improved accuracy. \rev{In standard SNNs, binary activation restricts the network's representation capacity, often leading to a severe accuracy degradation compared to ANNs. BuSNN uses graded multi-level spikes to enrich neuronal responses, thereby improving representational capacity and achieving stronger accuracy.}

Motivated by the discrete nature of spiking activations, we additionally construct activation-quantized ANNs for a more comprehensive comparison. The activations of ANNs are discretized with quantized representations. Specifically, we adopt asymmetric uniform quantization at 8-bit and 4-bit precision (INT8 with an integer range of $[0,255]$ and INT4 with $[0,15]$). For each layer, a scaling factor is estimated from activation quantiles to reduce the influence of outliers, after which floating-point activations are clipped and linearly mapped to the target integer range. This process makes the activation dynamics of ANNs comparable to the discrete behavior of SNNs. As shown in Table~\ref{tab:cl-imagenet}, our method outperforms the INT4 activation-quantized ANNs. Remarkably, BuSNN operates with $\theta_{\text{burst}} = 4$, providing only four discrete activation levels per neuron, which is fewer than the 16 levels in INT4 ANNs, yet it achieves higher accuracy. 

\rev{In addition to static image benchmarks, we further evaluate our model on the neuromorphic CIFAR10-DVS dataset~\cite{cifar10dvs}. 
Our method also achieves superior results, demonstrating its effectiveness beyond static image classification. 
Detailed results are provided in Supplementary Materials S3-\textit{A}.}

\subsection{Adversarial Attack Robustness}

We evaluate adversarial robustness under two classical methods, Fast Gradient Sign Method (FGSM)~\cite{goodfellow2014explaining} and Projected Gradient Descent (PGD)~\cite{madry2017towards}. 
FGSM applies a small perturbation along the sign of the input gradient to maximize the loss:
\begin{equation}
x_{\text{adv}} \;=\; x \;+\; \epsilon \cdot \operatorname{sign}\!\bigl(\nabla_x \mathcal{L}(\hat{y}(x), y)\bigr),
\end{equation}
where \(\epsilon\) controls the perturbation magnitude, \(\hat{y}(x)\) denotes the network output for input \(x\), \(y\) is the ground-truth label, and \(\nabla_x \mathcal{L}\) is the input gradient of the loss \(\mathcal{L}\). 
PGD is an iterative extension of FGSM that uses multiple small steps with projection:
\begin{equation}
x_{t+1} \;=\; \operatorname{Proj}_{\,x+\mathcal{S}}\!\left(x_t \;+\; \alpha \cdot \operatorname{sign}\!\bigl(\nabla_x \mathcal{L}(\hat{y}(x_t), y)\bigr)\right),
\end{equation}
where \(\alpha\) is the step size and \(t\) is the iteration index. The set \(\mathcal{S}=\{\delta:\|\delta\|_\infty \le \epsilon\}\) constrains the perturbation within \([-\epsilon,+\epsilon]\) in the \(\ell_\infty\) sense.


\rev{To facilitate robustness comparisons among models with different clean accuracies, we report both the absolute accuracy under adversarial perturbations and a normalized robustness metric, \emph{Score}. {It is} defined as $\text{Score}=A_{\mathrm{adv}}/A$, {which} measures the proportion of clean accuracy retained under adversarial attacks. Here \(A\) denotes the clean accuracy and \(A_{\mathrm{adv}}\) denotes the classification accuracy under FGSM or PGD attacks.
Equivalently, \emph{Score} can be viewed as one minus the relative accuracy degradation from clean to adversarially perturbed data.
A higher \emph{Score} indicates smaller relative degradation and thus stronger robustness. This metric provides a normalized view of robustness and complements absolute adversarial accuracy when comparing models with different clean accuracies.}

\rev{Table~\ref{tab:Adversarial_Attack} compares our method with advanced neuron models, as well as with several methods developed specifically for adversarial attack scenarios. We do not use adversarial training, since our goal is to evaluate intrinsic adversarial robustness rather than the effect of robust optimization. 
}


\rev{We report results for BuSNN and BuSNN- (BuSNN without DWC). On CIFAR-10, BuSNN achieves the best performance against single-step attacks such as FGSM. Under multi-iteration attacks, including PGD7 and PGD10, BuSNN also shows strong robustness among binary SNN models. It obtains the highest \emph{Score}s among the compared SNNs and remains close to the ANN baselines. On CIFAR-100, BuSNN outperforms ANNs under both clean and adversarial conditions. Its best \emph{Score}s further indicate stronger robustness.
}

\rev{This robustness gain comes from the complementary roles of BSNs and DWC.
BSNs produce graded burst responses.
When small membrane perturbations remain within the same response interval, the neuronal output does not change.
Even when perturbations cross response boundaries, the burst-count change is bounded by the interval structure, as discussed in Eq.~\eqref{eq:burst_stability}.
DWC further limits weight-side amplification by regularizing active connections and suppressing inactive connections with excessively large weights.
This constrains the weight magnitude term in Eq.~\eqref{eq:weight_amplification_bound}, thereby reducing the amplification of perturbation-induced activation changes.}


\begin{table*}
  \caption{Comparison of recognition accuracy (\%) of different methods under the Modified PLIF-Net architecture on CIFAR10-C. BuSNN- denotes BuSNN without DWC. \rev{For columns L1–L5, the leading value is averaged accuracy, and the value in parentheses denotes standard deviation across corruption types at each severity. To facilitate comparison, the reported standard deviation includes those of our method and the ANN baseline. \underline{Underline} indicates the optimal result.}}
  \centering
    \begin{tabularx}{\textwidth}{@{}l>{\hsize=1\hsize}X*{1}{X}{X}*{7}{X}@{}}
    \toprule
    & & \multicolumn{5}{c}{Severity Level}                   \\
    \cmidrule(r){3-7}
    Method   & Clean  & L1  & L2 & L3 & L4 & L5 & C-AVG & Score\\
    \midrule
    ANN   & 95.84 & 92.27 \textcolor{black}{(6.17)} & 89.65 \textcolor{black}{(6.53)} & 86.52 \textcolor{black}{(8.64)} & 81.22 \textcolor{black}{(13.11)} & 73.90 \textcolor{black}{(15.81)} & 84.71 & 0.8840 \\
    \midrule
    LIF \cite{lif}   & 95.04 & 90.90 & 88.05 & 84.85 & 79.32 & 71.50 & 82.92 & 0.8725    \\
    PLIF \cite{plif}  & 95.08 & 91.10 & 88.34 & 85.10 & 79.55 & 72.27 & 83.27 & 0.8758    \\
    NLIF \cite{nsnn}  & 94.69 & 90.22 & 86.52 & 82.37 & 76.25 & 66.95 & 80.46 & 0.8497  \\
    GLIF \cite{glif}  & 93.94 & 90.12 & 87.46 & 84.49 & 79.36 & 72.54 & 82.80 & 0.8814     \\
    SLTT \cite{sltt}  & 95.08 & 91.43 & 88.59 & 85.39 & 80.14 & 73.27 & 83.77 & 0.8810    \\
    LTF-ALSF \cite{fu2025adaptation}  & 94.48 & 88.54 & 83.76 & 79.18 & 73.22 & 64.57 & 77.85 & 0.8240 \\
    PSN \cite{psn}  & 95.32  & 91.68 & 88.90 & 85.79 & 80.75 & 73.28 & 84.08 & 0.8821 \\
    MPSN \cite{mpsn}  & 96.18 & 92.37 & 89.78 & 86.86 & 81.90 & 75.08 & 85.20 & 0.8858 \\
    BuSNN- (ours)  & \textbf{96.22} & \textbf{92.80} \textcolor{black}{(5.57)} & \textbf{90.27} \textcolor{black}{(6.51)} & \textbf{87.13} \textcolor{black}{(9.14)} & \textbf{82.07} \textcolor{black}{(13.80)} & \textbf{74.70} \textcolor{black}{(14.90)} & \textbf{85.39} & \textbf{0.8875} \\
    BuSNN (ours)  & \underline{\textbf{96.25}} & \underline{\textbf{93.04}} \textcolor{black}{(5.23)} & \underline{\textbf{90.64}} \textcolor{black}{(6.07)} & \underline{\textbf{87.59}} \textcolor{black}{(8.59)} & \underline{\textbf{82.75}} \textcolor{black}{(13.22)} & \underline{\textbf{75.74}} \textcolor{black}{(15.86)} & \underline{\textbf{85.95}} & \underline{\textbf{0.8930}} \\
    \bottomrule
  \end{tabularx}
  \label{tab:robust_cf}
\end{table*}

\begin{figure}[t]
  \centering
  \begin{subfigure}{0.53\linewidth}
    \centering
    \includegraphics[width=\linewidth]{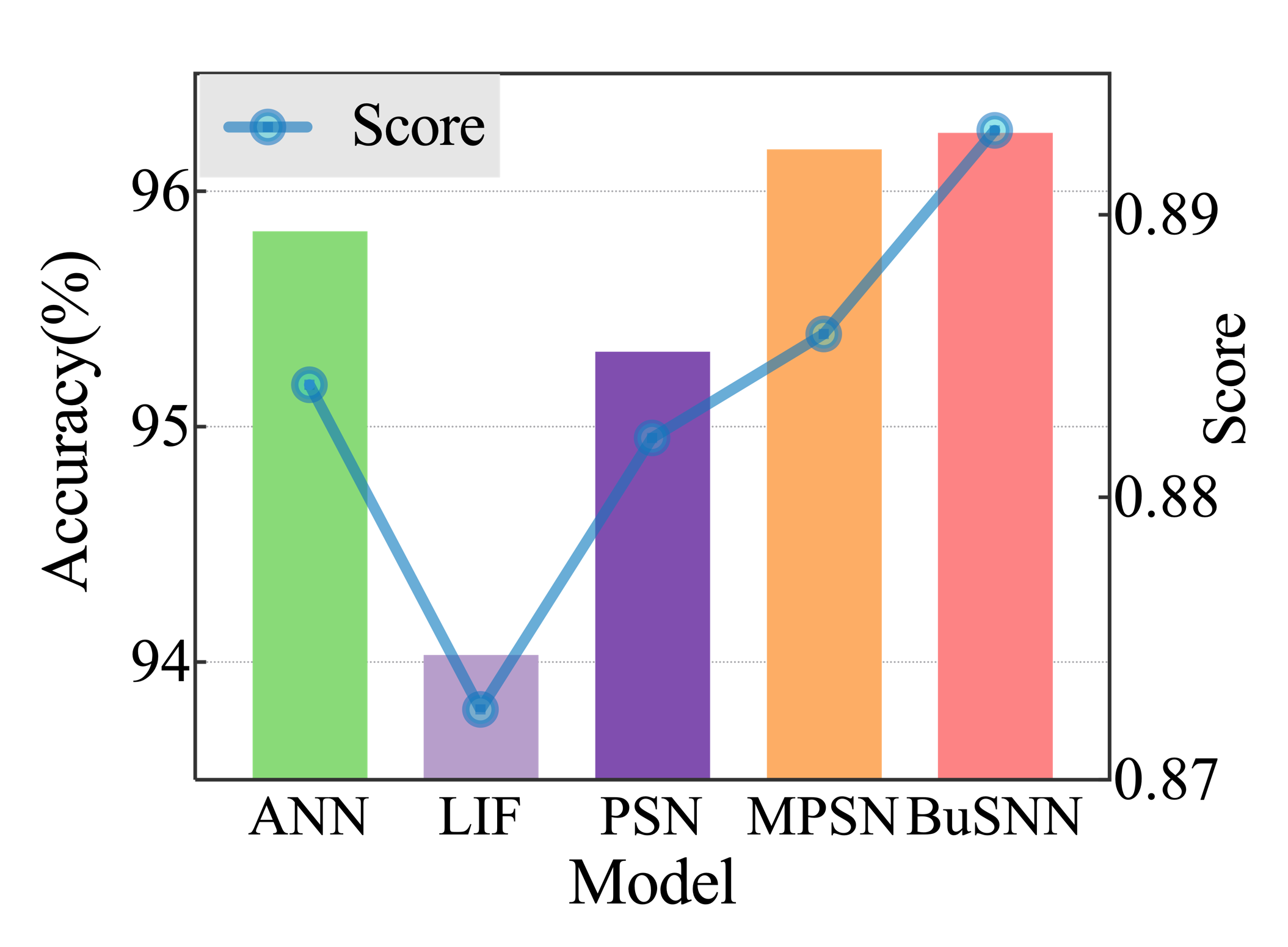}
    \caption{Clean}
    \label{}
  \end{subfigure}\hfill%
  \begin{subfigure}{0.47\linewidth}
    \centering
    \includegraphics[width=\linewidth]{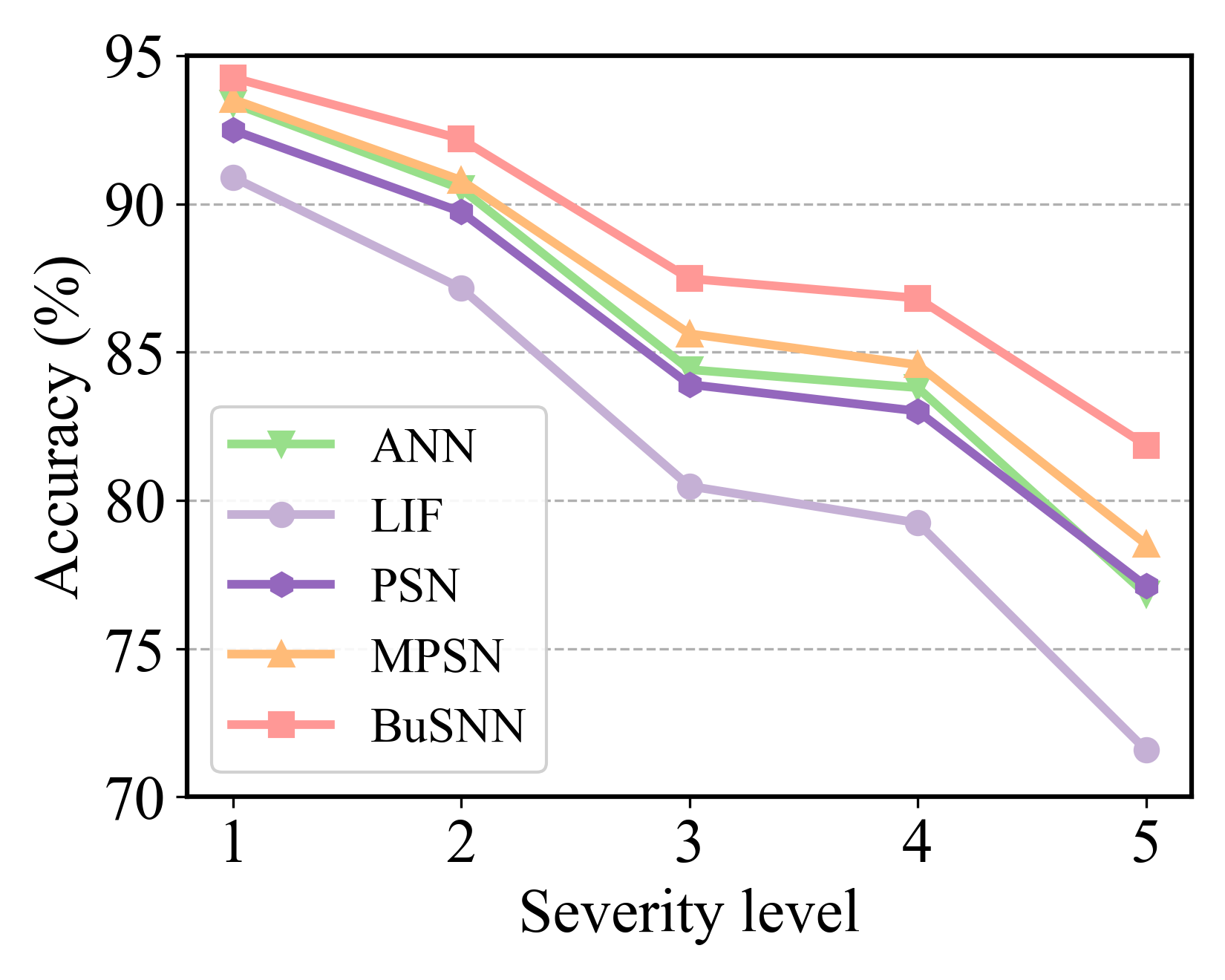}
    \caption{Glass blur}
    \label{}
  \end{subfigure}

  \vspace{0.6em}

  \begin{subfigure}{0.47\linewidth}
    \centering
    \includegraphics[width=\linewidth]{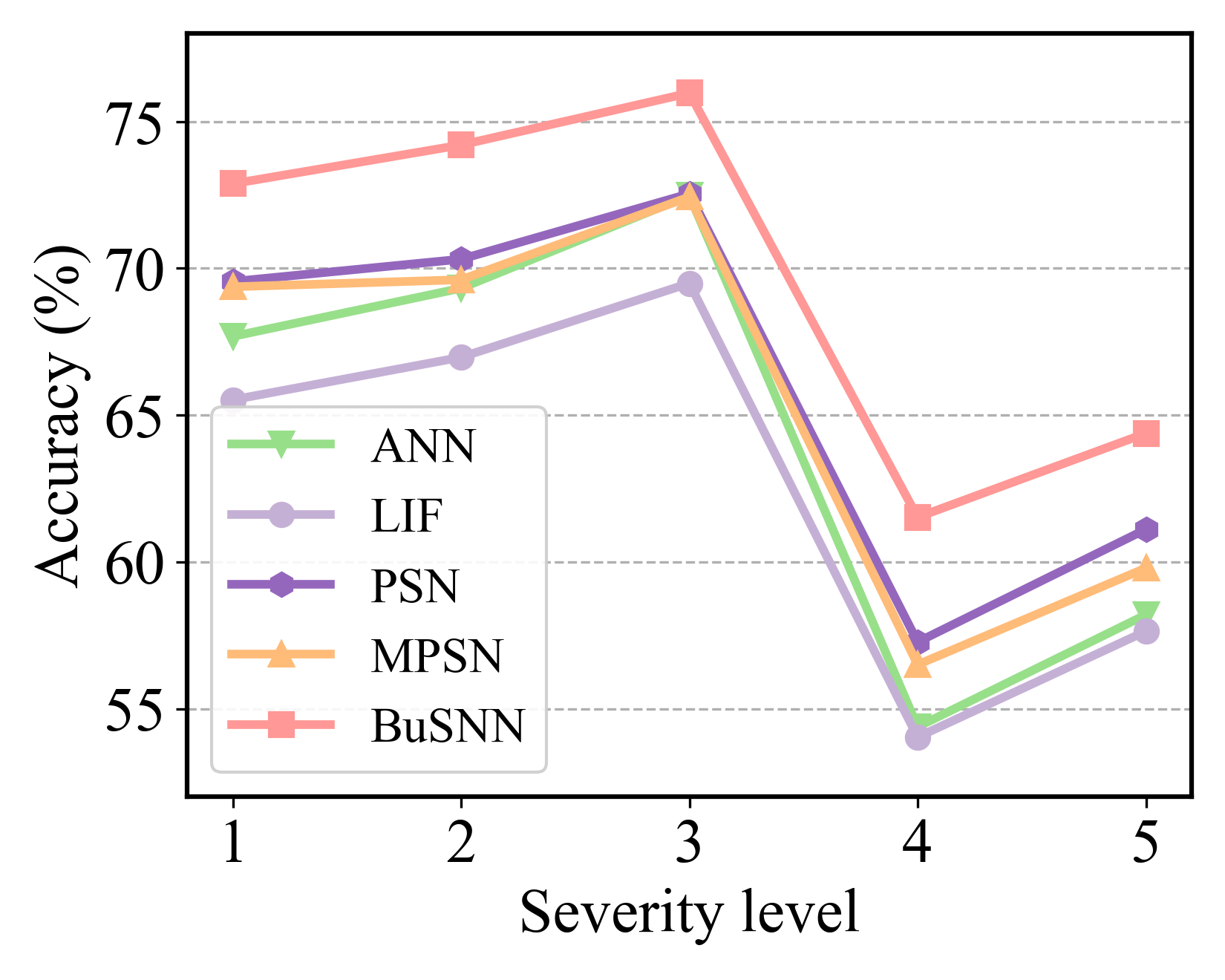}
    \caption{Frost}
    \label{}
  \end{subfigure}\hfill
  \begin{subfigure}{0.47\linewidth}
    \centering
    \includegraphics[width=\linewidth]{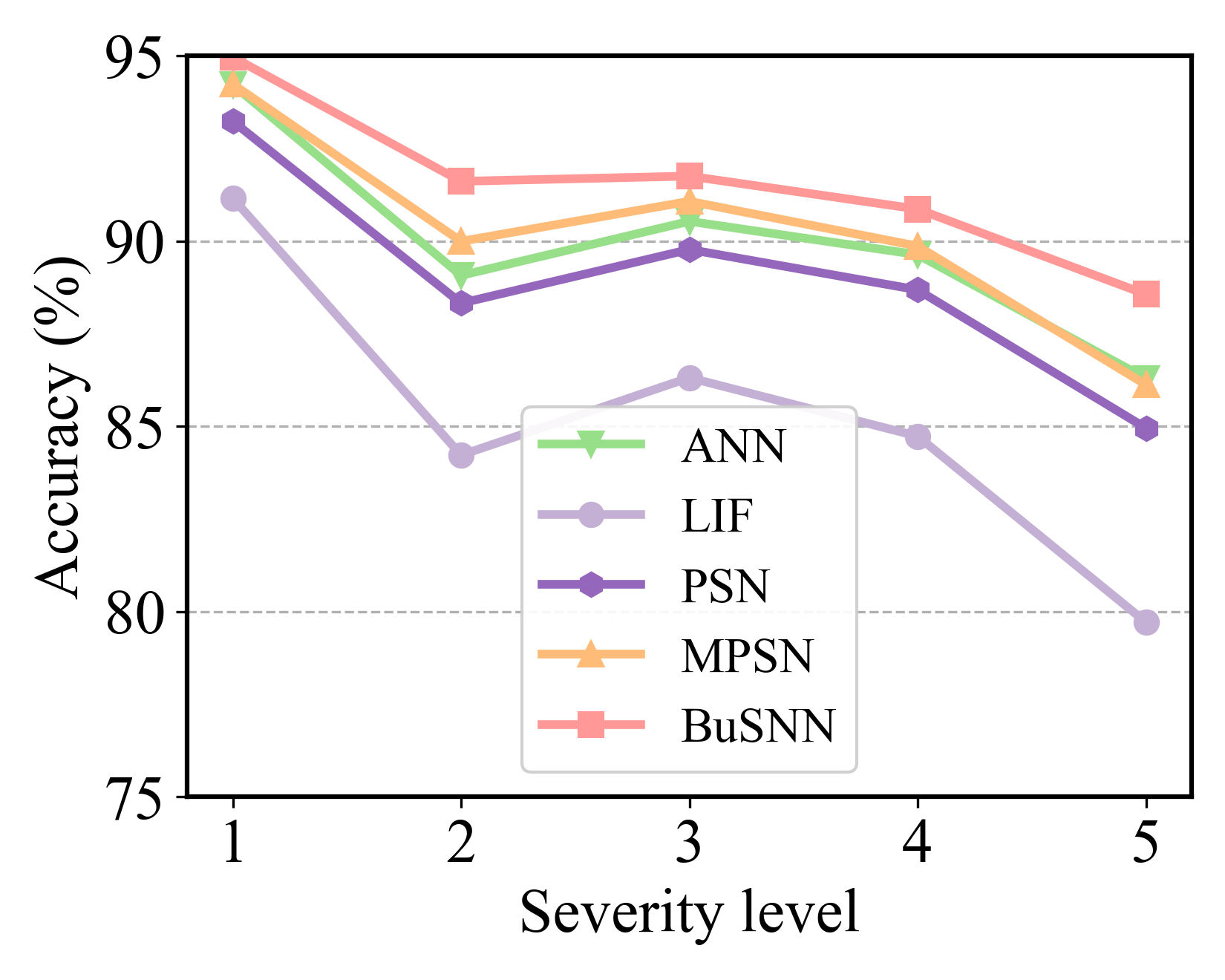}
    \caption{Snow}
    \label{}
  \end{subfigure}
  \caption{\rev{Comparison of clean accuracy and corruption robustness \emph{Score} on CIFAR-10. (a) The bars correspond to the left (y)-axis and represent clean test accuracy, while the line corresponds to the right (y)-axis and represents the normalized corruption robustness \emph{Score}. (b--d) Accuracy on CIFAR-10-C under glass blur, frost, and snow corruptions across severity levels L1 through L5.}}
  \label{fig:acc_example}
\end{figure}

\begin{table*}
  \caption{Comparison of recognition accuracy (\%) of different methods on IMAGENET-C. BuSNN- denotes the BuSNN without using DWC. \rev{Columns L1–L5 report the averaged accuracy at each severity level, with the value in parentheses denoting the standard deviation across corruption types. C-AVG is the average over all severities and corruption types. \underline{Underlined} results denote the best performance among compared SNNs. Results in this table are obtained from our reproduced experiments.}}
  \centering
    \begin{tabularx}{\textwidth}{@{}l p{2.1cm} p{1.1cm} *{5}{X} p{1.2cm} X@{}}
    \toprule
    & & & \multicolumn{5}{c}{Severity Level}                   \\
    \cmidrule(r){4-8}
    Method  & Spiking Network & Clean  & L1  & L2 & L3 & L4 & L5 & C-AVG & Score\\
    \midrule
    ANN \cite{ResNet18} & ResNet-18  & 69.76 & 54.25\textcolor{black}{(7.93)} & 43.82\textcolor{black}{(11.32)} & 34.55\textcolor{black}{(14.78)} & 23.93\textcolor{black}{(14.84)} & 15.50\textcolor{black}{(12.75)} & 34.41 & 0.4933\\
    ANN-INT8  & ResNet-18  & 68.05 & 49.30\textcolor{black}{(7.56)} & 38.33\textcolor{black}{(11.05)} & 29.49\textcolor{black}{(14.24)} & 20.26\textcolor{black}{(13.99)} & 12.97\textcolor{black}{(12.00)} & 30.07 & 0.4419\\
    ANN-INT4  & ResNet-18  & 66.44 & 47.10\textcolor{black}{(7.60)} & 35.86\textcolor{black}{(10.80)} & 26.92\textcolor{black}{(14.04)} & 17.97\textcolor{black}{(13.34)} & 11.31\textcolor{black}{(10.83)} & 27.83 & 0.4189\\
    \cmidrule(r){2-10}
    MS \cite{hu2024advancing} & MS ResNet-18  & 63.10 & 43.62 & 32.38 & 23.84 & 15.76 & 9.49 & 25.02 & 0.3965 \\
    BuSNN (ours)  & MS ResNet-18 & \textbf{69.64} & \textbf{48.67}\textcolor{black}{(7.82)} & \textbf{36.64}\textcolor{black}{(10.99)} & \textbf{26.75}\textcolor{black}{(13.85)} & \textbf{17.63}\textcolor{black}{(13.44)} & \textbf{11.00}\textcolor{black}{(11.52)} & \textbf{28.14} & \textbf{0.4040} \\
    \cmidrule(r){2-10}  
    SEW \cite{sew} & SEW ResNet-18  & 63.18 & 41.19 & 30.45 & 23.22 & 16.24 & 10.58 & 24.34 & 0.3853 \\
    PLIF \cite{plif} & SEW ResNet-18 & 66.19 & 45.66 & 34.61 & 26.22 & 17.77 & 11.03 & 27.06 & 0.4088  \\
    NLIF \cite{nsnn}  & SEW ResNet-18  & 63.32 & 45.38 & 34.31 & 26.16 & 17.87 & 11.26 & 27.00 & 0.4264 \\ 
    PSN \cite{psn} & SEW ResNet-18  & 67.63 & 47.83 & 36.79 & 28.16 & 19.13 & 12.05 & 28.79 & 0.4257 \\
    MPSN \cite{mpsn} & SEW ResNet-18 & 67.92 & 48.25 & 37.16 & 28.50 & 19.39 & 12.29 & 29.12 & 0.4287 \\
    BuSNN- (ours) & SEW ResNet-18 & \textbf{69.89} & \textbf{51.05}\textcolor{black}{(7.08)} & \textbf{40.00}\textcolor{black}{(10.06)} & \textbf{30.91}\textcolor{black}{(13.76)} & \textbf{21.19}\textcolor{black}{(13.60)} & \textbf{13.47}\textcolor{black}{(11.33)} & \textbf{31.32} & \textbf{0.4482} \\
    BuSNN (ours)  & SEW ResNet-18 & \underline{\textbf{69.92}} & \underline{\textbf{51.29}}\textcolor{black}{(6.86)} & \underline{\textbf{40.28}}\textcolor{black}{(9.85)} & \underline{\textbf{31.28}}\textcolor{black}{(13.50)} & \underline{\textbf{21.45}}\textcolor{black}{(13.46)} & \underline{\textbf{13.60}}\textcolor{black}{(11.28)} & \underline{\textbf{31.58}} & \underline{\textbf{0.4517}} \\

    \cmidrule(r){1-10}    
    \cmidrule(r){1-10}    ANN \cite{ResNet18} & ResNet-34  & 73.31 & 59.75\textcolor{black}{(6.58)} & 49.57\textcolor{black}{(10.49)} & 40.39\textcolor{black}{(14.36)} & 29.29\textcolor{black}{(15.68)} & 19.70\textcolor{black}{(14.44)} & 39.74 & 0.5421 \\
    ANN-INT8  & ResNet-34  & 71.50 & 54.82\textcolor{black}{(6.39)} & 44.23\textcolor{black}{(10.26)} & 34.94\textcolor{black}{(13.67)} & 24.56\textcolor{black}{(14.47)} & 15.95\textcolor{black}{(13.34)} & 34.90 & 0.4881 \\
    ANN-INT4  & ResNet-34  & 69.73 & 52.54\textcolor{black}{(6.53)} & 41.56\textcolor{black}{(10.15)} & 32.10\textcolor{black}{(13.61)} & 21.76\textcolor{black}{(14.04)} & 13.63\textcolor{black}{(12.37)} & 32.32 & 0.4635\\
    \cmidrule(r){2-10}
    MS \cite{hu2024advancing} & MS ResNet-34  & 69.65 & 51.87 & 40.53 & 31.16 & 21.39 & 13.62 & 31.71 & 0.4553 \\
    BuSNN (ours) & MS ResNet-34  & \underline{\textbf{72.60}} & \textbf{55.11}\textcolor{black}{(6.51)} & \textbf{43.78}\textcolor{black}{(10.20)} & \textbf{34.17}\textcolor{black}{(14.11)} & \textbf{23.59}\textcolor{black}{(14.62)} & \textbf{15.22}\textcolor{black}{(12.51)} & \textbf{34.37} & \textbf{0.4734} \\
    \cmidrule(r){2-10}  
    SLTT \cite{sltt} &  ResNet-34 & 66.19 & 45.76 & 35.77 & 28.10 & 19.50 & 12.67 & 28.36 & 0.4285\\
    SEW \cite{sew} & SEW ResNet-34  & 67.04 & 46.38 & 35.34 & 27.05 & 18.90 & 12.32 & 28.00 & 0.4177 \\
    PSN \cite{psn} & SEW ResNet-34  & 70.18 & 52.12 & 41.44 & 32.96 & 23.13 & 14.82 & 32.89 & 0.4687 \\
    MPSN \cite{mpsn} & SEW ResNet-34 & 71.30 & 53.67 & 43.03 & 34.38 & 24.23 & 15.58 & 34.18 & 0.4794 \\
    BuSNN- (ours) & SEW ResNet-34 & \textbf{72.23} & \textbf{55.05}\textcolor{black}{(6.23)} & \textbf{44.74}\textcolor{black}{(9.52)} & \textbf{35.88}\textcolor{black}{(12.93)} & \textbf{25.47}\textcolor{black}{(13.55)} & \textbf{16.44}\textcolor{black}{(12.35)} & \textbf{35.52} & \textbf{0.4918} \\
    BuSNN (ours)  & SEW ResNet-34 & \textbf{72.40} & \underline{\textbf{55.91}}\textcolor{black}{(6.13)} & \underline{\textbf{45.68}}\textcolor{black}{(9.35)} & \underline{\textbf{36.88}}\textcolor{black}{(12.77)} & \underline{\textbf{26.22}}\textcolor{black}{(13.54)} & \underline{\textbf{16.91}}\textcolor{black}{(12.45)} &  \underline{\textbf{36.32}} & \underline{\textbf{0.5017}}\\

    \cmidrule(r){1-10}  
    ANN \cite{ResNet18} & ResNet-50  & 76.13 & 61.89\textcolor{black}{(6.52)} & 51.59\textcolor{black}{(10.47)} & 41.60\textcolor{black}{(14.44)} & 29.72\textcolor{black}{(15.72)} & 19.44\textcolor{black}{(14.23)} & 40.85 & 0.5365 \\
    ANN-INT8 & ResNet-50   & 74.54 & 56.91\textcolor{black}{(6.67)} & 45.87\textcolor{black}{(10.59)} & 36.13\textcolor{black}{(14.24)} & 25.44\textcolor{black}{(14.82)} & 16.24\textcolor{black}{(13.24)} & 36.12 & 0.4846 \\
    ANN-INT4 & ResNet-50  & 72.10  & 50.39\textcolor{black}{(6.93)} & 39.39\textcolor{black}{(10.17)} & 30.28\textcolor{black}{(14.03)} & 20.79\textcolor{black}{(14.06)} & 13.14\textcolor{black}{(11.84)} & 30.80  & 0.4272 \\
    \cmidrule(r){2-10}  
    MS \cite{hu2024advancing} & MS ResNet-50  & 71.05 & 52.10 & 40.48 & 30.77 & 21.18 & 13.35 & 31.58 & 0.4444 \\
    BuSNN (ours) & MS ResNet-50 & \underline{\textbf{73.70}} & \underline{\textbf{54.57}}\textcolor{black}{(7.40)} &\underline{\textbf{42.76}}\textcolor{black}{(11.32)} & \underline{\textbf{32.92}}\textcolor{black}{(14.94)} & \underline{\textbf{22.59}}\textcolor{black}{(14.52)} & \underline{\textbf{14.26}}\textcolor{black}{(11.83)} & \underline{\textbf{33.42}} & \underline{\textbf{0.4535}} \\ 
    \cmidrule(r){2-10}  
    SLTT \cite{sltt} & ResNet-50 & 67.02 & 45.99 & 35.36 & 27.41 & 19.12 & 12.43 & 28.06 & 0.4187 \\
    SEW \cite{sew} & SEW ResNet-50  & 67.57 & 44.77 & 33.36 & 25.44 & 17.81 & 11.44 & 26.56 & 0.3931  \\
    PSN \cite{psn} & SEW ResNet-50 & 70.75 & 49.42 & 37.80 & 28.61 & 19.72 & 12.36 & 29.58 & 0.4181 \\
    MPSN \cite{mpsn} & SEW ResNet-50 & 71.56 & 51.21 & 39.52 & 30.24 & 20.80 & 13.06 & 30.97 & 0.4327 \\
    BuSNN- (ours) & SEW ResNet-50 & \textbf{72.59} & \textbf{52.34}\textcolor{black}{(7.40)} & \textbf{41.04}\textcolor{black}{(10.57)} & \textbf{31.66}\textcolor{black}{(14.01)} & \textbf{22.08}\textcolor{black}{(13.84)} & \textbf{14.09}\textcolor{black}{(11.69)} & \textbf{32.24} & \textbf{0.4442} \\
    BuSNN (ours) & SEW ResNet-50 & \textbf{72.70} & \textbf{52.61}\textcolor{black}{(7.66)} & \textbf{41.25}\textcolor{black}{(10.86)} & \textbf{31.84}\textcolor{black}{(14.17)} & \textbf{22.28}\textcolor{black}{(13.92)} & \textbf{14.23}\textcolor{black}{(11.65)} & \textbf{32.44} & \textbf{0.4462} \\

    \bottomrule
  \end{tabularx}
  \label{tab:robust_im}
\end{table*}

\begin{table}[t]
\caption{Comparison of computational cost, energy consumption, and performance for different models on CIFAR-10 and ImageNet. \rev{Results are reported using Modified PLIF-Net on CIFAR-10 ($32\times32$) and SEW-ResNet-18 on ImageNet ($224\times224$).}}
\centering
\begin{tabular*}{0.48\textwidth}{@{\extracolsep{\fill}} lccccc}
\toprule
{Model} & {AC (M)} & {MAC (M)} & {Energy (mJ)} & \textcolor{black}{Clean} & \textcolor{black}{C-AVG} \\
\midrule
\multicolumn{6}{c}{\textcolor{black}{\textit{CIFAR-10 $(32\times32)$}}} \\
\hline
LIF &604.559 & 28.312 & 0.67 & \textcolor{black}{95.04} & \textcolor{black}{82.92} \\
PSN & 1310.528 & 28.312 & 1.31 & \textcolor{black}{95.32} & \textcolor{black}{84.08} \\
MPSN & 1687.701 & 28.312 & 1.65 & \textcolor{black}{96.18} & \textcolor{black}{85.20} \\
ReLU & - & 1735.172 & 7.98 & \textcolor{black}{95.84} & \textcolor{black}{84.71}  \\
BuSNN & 1879.055 & 28.312 & 1.82 & \textcolor{black}{96.25} & \textcolor{black}{85.95}  \\
\midrule
\multicolumn{6}{c}{\textcolor{black}{\textit{ImageNet $(224\times224)$}}} \\
\hline
\textcolor{black}{SEW} & \textcolor{black}{908.25} & \textcolor{black}{472.056} & \textcolor{black}{2.99} & \textcolor{black}{63.18} & \textcolor{black}{24.34} \\
\textcolor{black}{PSN} & \textcolor{black}{2148.29} & \textcolor{black}{472.056} & \textcolor{black}{4.10} & \textcolor{black}{67.63} & \textcolor{black}{28.79} \\
\textcolor{black}{MPSN} & \textcolor{black}{2041.90} & \textcolor{black}{472.056} & \textcolor{black}{4.01} & \textcolor{black}{67.92} & \textcolor{black}{28.79} \\
\textcolor{black}{ReLU} & \textcolor{black}{-} & \textcolor{black}{1817.310} & \textcolor{black}{8.36} & \textcolor{black}{69.76} & \textcolor{black}{34.41} \\
\textcolor{black}{BuSNN} & \textcolor{black}{1358.116} & \textcolor{black}{472.056} & \textcolor{black}{3.39} & \textcolor{black}{69.92} & \textcolor{black}{31.58}  \\
\bottomrule
\end{tabular*}
\label{tab:comput}
\end{table}

\subsection{Common Corruption Robustness}

Robustness to common corruptions is an essential indicator of a model’s generalization ability under image conditions. To assess this property, we adopt the CIFAR10-C~\cite{hendrycks2019benchmarking} and IMAGENET-C~\cite{hendrycks2019benchmarking} benchmarks, which include 19 types of image corruptions, each applied at five levels of severity. Representative examples are shown in Fig.~\ref{fig:fig1} (a). All models are trained solely on clean data and are directly evaluated on the corrupted datasets without any additional fine-tuning.

\rev{We use the same normalized \emph{Score} defined above to evaluate corruption robustness, with \(A_{\mathrm{adv}}\) replaced by the corrupted-data accuracy \(A_c\). Accordingly, the corruption robustness \emph{Score} is computed as $\text{Score}=A_c/A$, where \(A\) denotes the clean accuracy. For each corruption type, \(A_c\) is averaged over the five severity levels as $A_c=\frac{1}{5}\sum_{l=1}^{5}A_l$, where \(A_l\) denotes the accuracy at severity level \(l\).}

On CIFAR10-C, Table~\ref{tab:robust_cf} summarizes accuracy under the Modified PLIF-Net backbone. Our BuSNN consistently improves average accuracy across all severities and outperforms prior SNNs. \rev{Notably, it maintains higher accuracy under conditions of greater corruption severity, such as levels \(L4\) and \(L5\).} \rev{BuSNN presents minor standard deviations for different corruption types at each severity, indicating that it maintains stable performance against various corruption types. Meanwhile, BuSNN also achieves the highest \emph{Score}, confirming that its robustness gains are not merely due to higher clean accuracy.} Compared with the ANN trained under the same conditions, BuSNN also demonstrates superior performance. We also present comparative results under individual corruption types in Fig.~\ref{fig:acc_example}.
\rev{The comparison between MPSN and BuSNN in Fig.~\ref{fig:acc_example}(a) shows that similar clean accuracy can still correspond to different corruption robustness \emph{Scores}. BuSNN combines the highest correspond to robustness \emph{Score} with the highest clean accuracy, demonstrating that stronger robustness is achieved while preserving clean performance. Across the three representative corruptions in Fig.~\ref{fig:acc_example}(b)--(d), BuSNN consistently maintains the highest accuracy on CIFAR10-C at all severity levels. This advantage persists as corruption strength varies, rather than arising from a single severity level. A detailed breakdown of the results by corruption type is provided in Supplementary Materials S3-\textit{B}.}

As shown in Table~\ref{tab:robust_im}, on the more challenging IMAGENET-C benchmark, our method remains competitive and establishes a new state of the art among SNNs.
We also compare SNNs with conventional ANNs under matched backbones. Full-precision ANNs (ResNet-18/34/50) still reach the highest robustness on IMAGENET-C. However, our method reduces the gap substantially and even surpasses quantized ANN baselines. With SEW ResNet-18 and SEW ResNet-34, BuSNN produces higher C-AVG than both ANN-INT8 and ANN-INT4, whereas no previous SNN method surpasses ANN-INT8 under the same settings. In terms of \emph{Score}, BuSNN delivers consistent gains over all previous SNNs at every severity level, indicating better corruption robustness and clean accuracy. Results obtained with MS ResNet backbones further demonstrate the generality of the proposed BuSNN, showing that it can effectively enhance both performance and robustness across diverse spiking residual network architectures. \rev{A detailed discussion of different corruption types can be found in Supplementary Materials S3-\textit{B}.}

\rev{The comparison with MPSN shows that BSNs provide richer and more stable neuronal representations for large-scale corrupted inputs. The further improvement of BuSNN over BuSNN- indicates that DWC additionally strengthens robustness by constraining synaptic weights under severe corruptions. \rev{This observation is consistent with the robustness analysis in Section \ref{method:D.3}}.}

\subsection{Computation Overhead}

\rev{SNNs have long been considered to have energy-efficiency advantages. 
To examine whether BuSNN preserves this desirable property, we conduct a theoretical analysis of computational overhead. 
Following the methodology established in~\cite{yao2025scaling}, we estimate the overhead of different methods in terms of Accumulation (AC) operations, Multiply-Accumulate (MAC) operations, and theoretical total energy consumption.}

\rev{The estimation of AC and MAC operations depends on the computational paradigms of different network models. 
In SNNs, each presynaptic spike triggers the accumulation of the corresponding synaptic weight into the postsynaptic membrane potential, which is counted as an AC operation. 
Accordingly, the AC operations reported in Table~\ref{tab:comput} are estimated by scaling the layer-wise synaptic operations with the spiking activity measured during inference over all time steps, with the burst count in BuSNN treated as accumulated spike events. 
By contrast, standard ReLU-based ANNs rely on continuous-valued dense activations, so their feature extraction layers are dominated by MAC operations. 
Thus, standalone AC operations are not considered in this theoretical evaluation. 
The total energy is then estimated using the corresponding unit energy costs following~\cite{yao2025scaling}, with detailed calculation and implementation details provided in Supplementary Materials S4.}

\rev{Based on these estimates, Table~\ref{tab:comput} shows that BuSNN achieves a consistent energy-performance trade-off on both CIFAR-10 and ImageNet. 
Compared with ReLU-based ANN baselines, BuSNN substantially reduces energy consumption through AC-dominated spiking computation. 
It achieves a \(4.38\times\) energy reduction on CIFAR-10 (1.82 mJ vs. 7.98 mJ) and a \(2.47\times\) reduction on ImageNet (3.39 mJ vs. 8.36 mJ). 
Compared with other SNN models such as LIF or SEW, BuSNN introduces a moderate increase in AC operations and energy cost. 
This additional cost is accompanied by stronger accuracy and robustness. 
These results indicate that BuSNN preserves the energy advantage of SNNs while achieving a better balance between efficiency, accuracy, and robustness.}

Furthermore, previous work has demonstrated that similar theoretical energy efficiency gains successfully translate to practical hardware implementations~\cite{yao2024spikenc,s21134462}. By applying the BuSNN framework in future chip designs, we expect to achieve not only low power consumption but also the high accuracy and robustness demonstrated in our experiments, thereby advancing the practical deployment of SNNs in real-world applications.

\subsection{Ablation Study} \label{sec:ablation}

\begin{table}[t]
\centering
\caption{Ablation study of burst spiking in BuSNN on the CIFAR-10 dataset. 
C-AVG and  \emph{Score} denote robustness results on CIFAR10-C, and FGSM denotes adversarial accuracy under FGSM attack.}
\label{tab:ablation_burst}
\begin{tabular}{lcccc}
\toprule
{Method} & {Clean} & {C-AVG} & {Score} & {FGSM} \\
\midrule
Binary spiking & 96.08 & 84.84 & 0.8830 & 63.08 \\
Burst spiking & \textbf{96.25} & \textbf{85.95} & \textbf{0.8930} & \textbf{67.89} \\
\bottomrule
\end{tabular}
\end{table}
\subsubsection{Burst spiking} \label{sec:burst}

We investigate the effect of burst spiking on SNN performance. 
When the maximum number of spikes per burst is set to one, the proposed burst mechanism degenerates into the binary spiking scheme. 
Table~\ref{tab:ablation_burst} compares BuSNN with burst spiking and binary spiking neurons on the CIFAR-10 dataset using the Modified PLIF-Net architecture.
The results show that burst spiking consistently improves both clean accuracy and robustness under common corruption and adversarial settings.
This improvement reflects enhanced representational capacity and increased network stability against input perturbations. Additional results for different settings of the maximum number of spikes per burst are provided in Supplementary Materials \rev{S5}.

\subsubsection{DWC}

As summarized in the preceding tables, we report results for both BuSNN- and BuSNN, where BuSNN- denotes BuSNN without using DWC. Based on Tables~\ref{tab:cl-cifar} and~\ref{tab:cl-imagenet}, BuSNN consistently achieves higher accuracy across datasets and architectures. 
Furthermore, Tables~\ref{tab:Adversarial_Attack},~\ref{tab:robust_cf}, and~\ref{tab:robust_im} show that incorporating DWC improves model robustness in most cases. 
For example, DWC raises the FGSM adversarial robustness accuracy of BuSNN- on CIFAR-100 from 20.84\% to 29.90\%, and enhances the corruption robustness \emph{Score} on IMAGENET-C using the SEW ResNet-18 backbone from 0.4482 to 0.4517.
These results demonstrate that DWC can enhance robustness against both common corruptions and adversarial perturbations without compromising clean accuracy.

\begin{figure}[t]
  \centering
  \begin{subfigure}{0.495\linewidth}
    \centering
    \includegraphics[width=\linewidth]{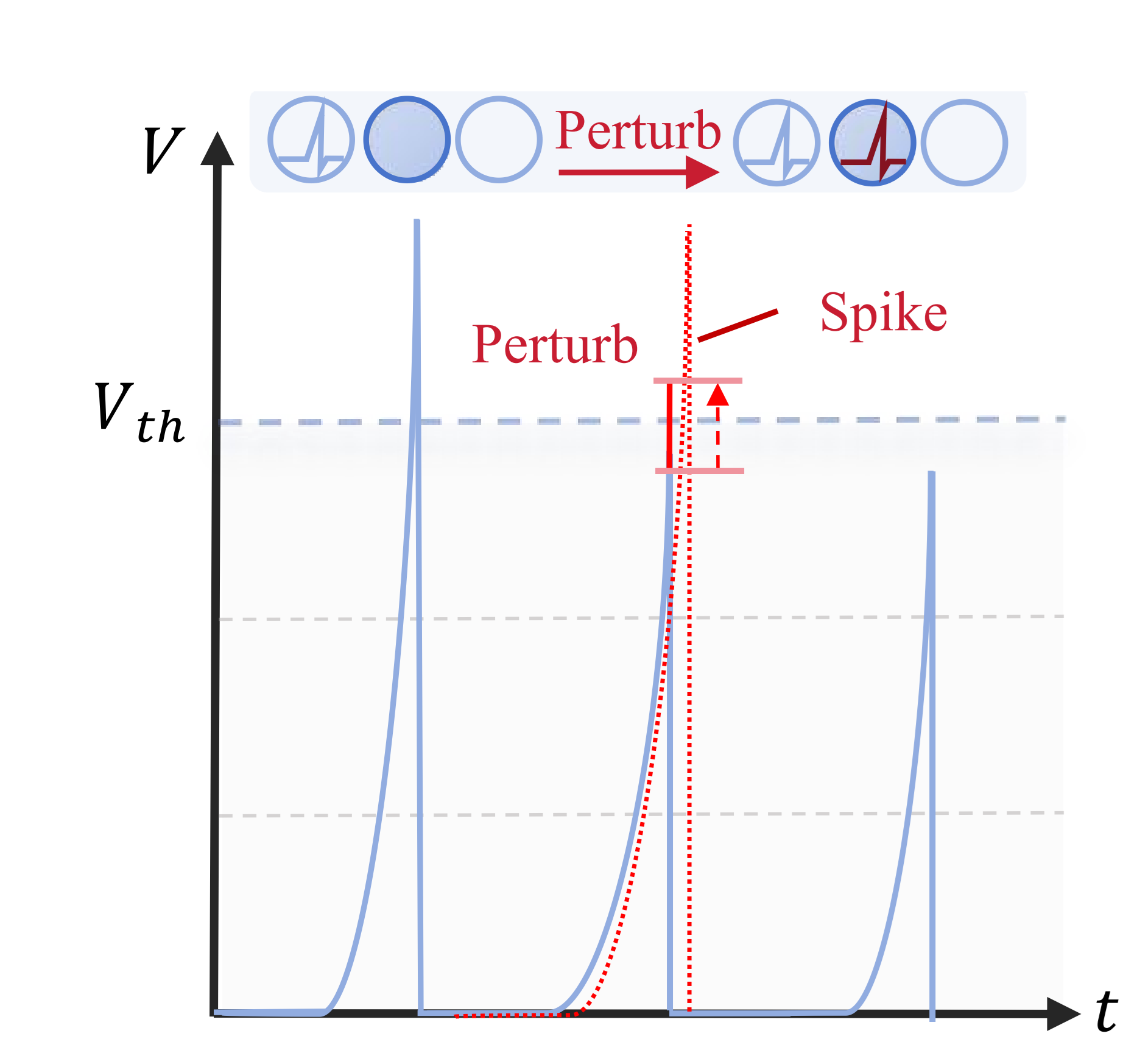}
    \caption{Binary spiking}
    \label{fig9_a}
  \end{subfigure}\hfill%
  \begin{subfigure}{0.495\linewidth}
    \centering
    \includegraphics[width=\linewidth]{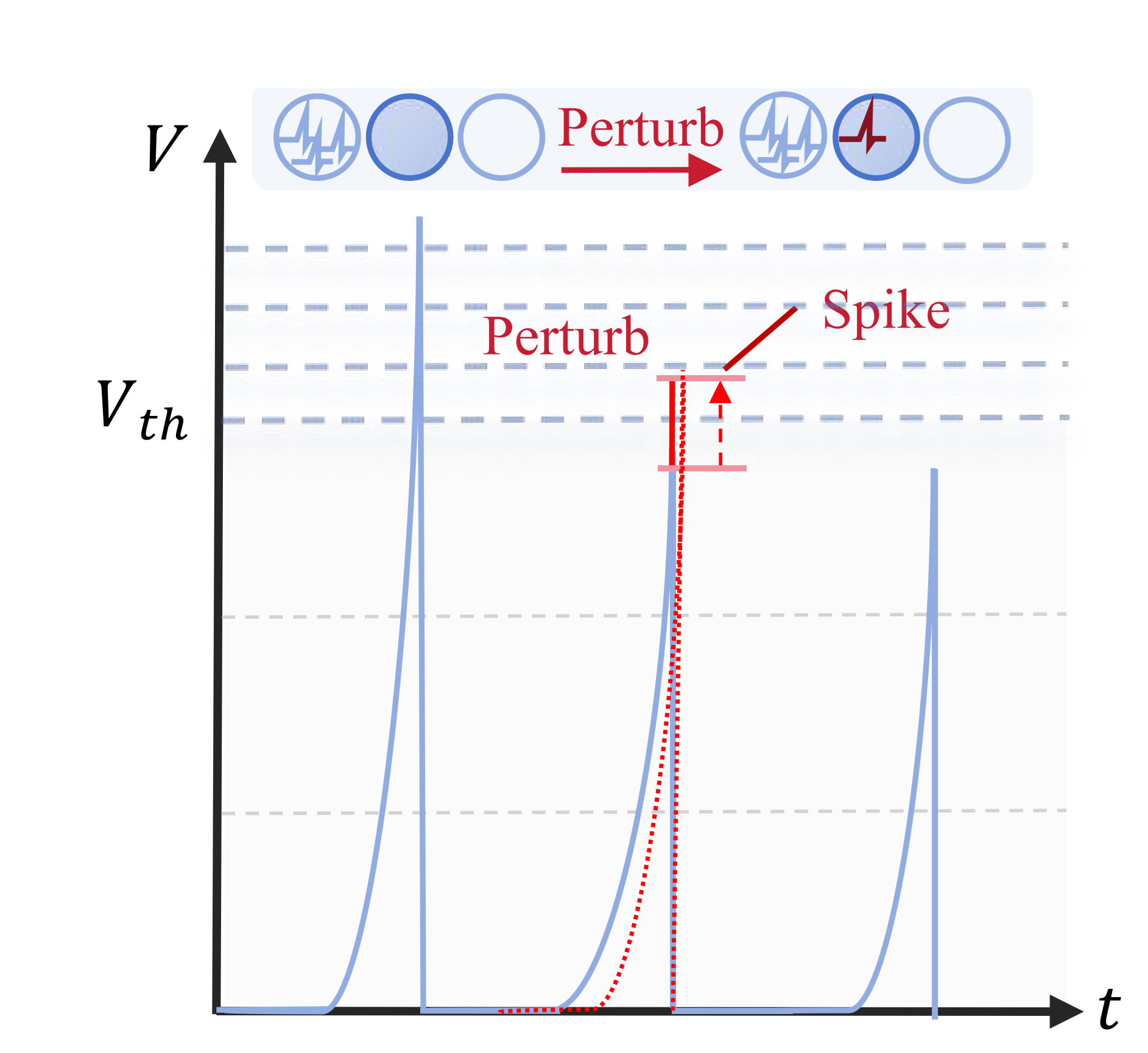}
    \caption{Burst spiking}
    \label{fig9_b}
  \end{subfigure}
  \begin{subfigure}{0.495\linewidth}
    \centering
    \includegraphics[width=\linewidth]{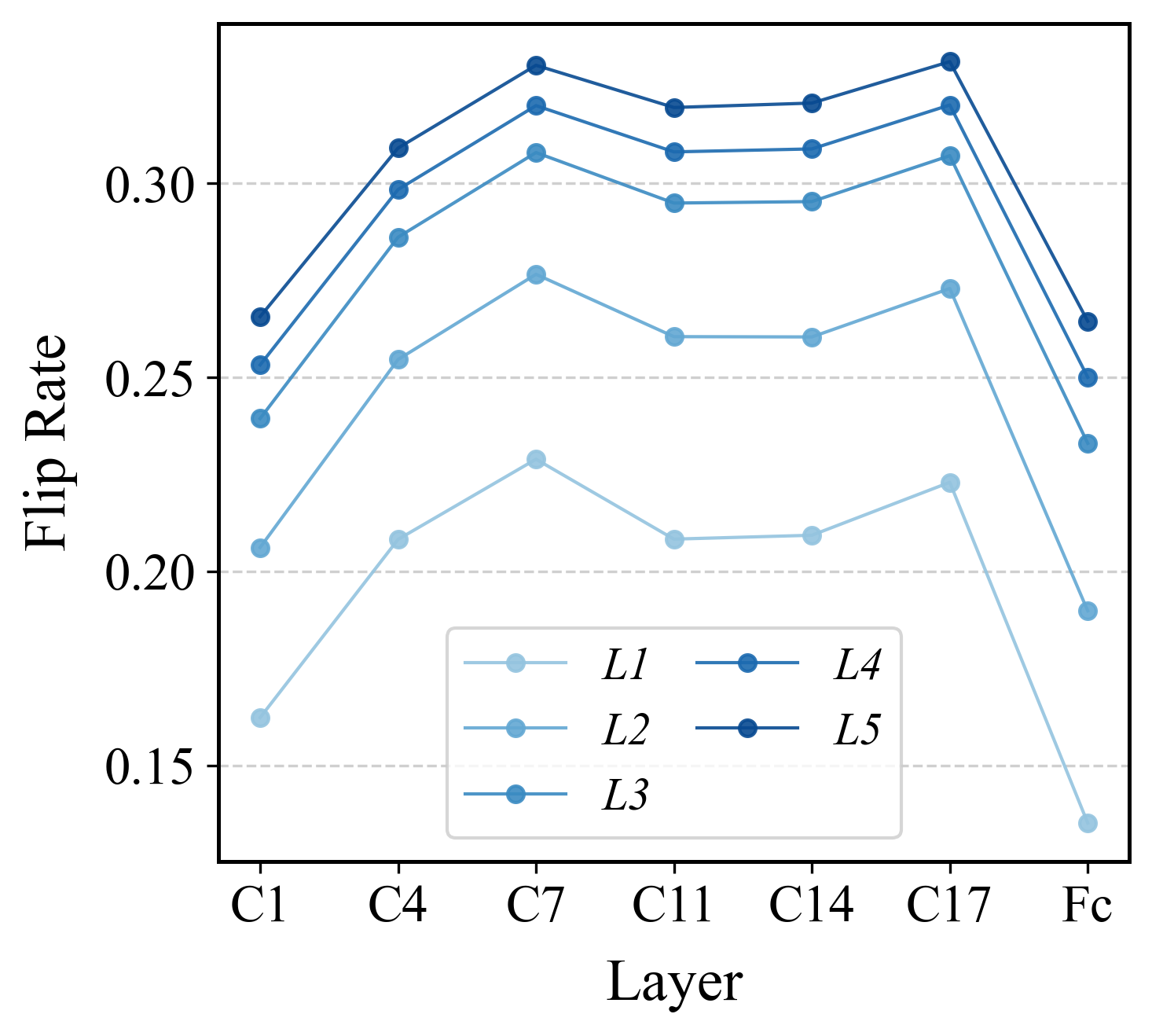}
    \caption{Flip rate of binary spiking}
    \label{fig9_c}
  \end{subfigure}\hfill%
    \begin{subfigure}{0.495\linewidth}
    \centering
    \includegraphics[width=\linewidth]{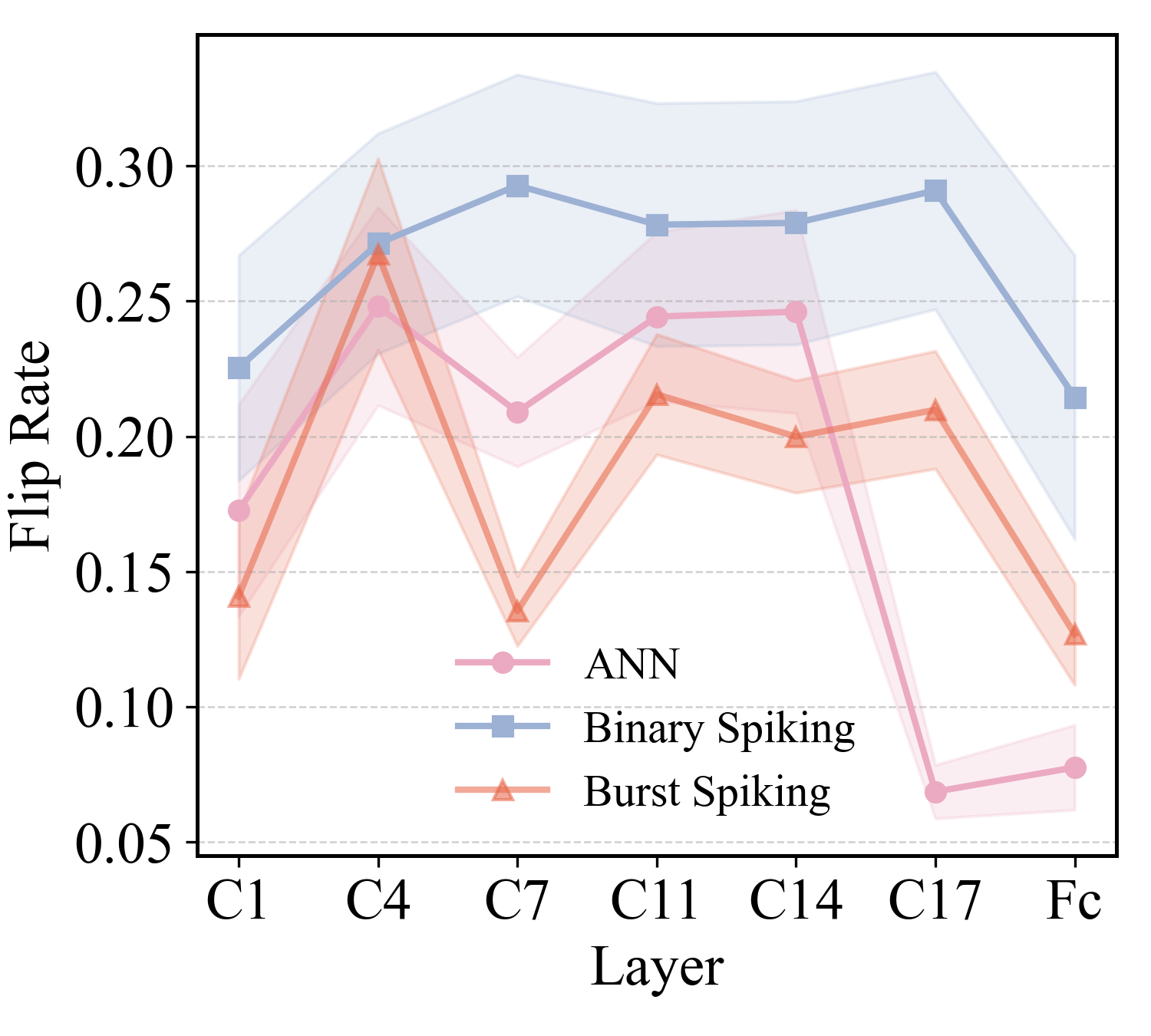}
    \caption{Flip rate comparison}
    \label{fig9_d}
  \end{subfigure}
  
  \caption{Comparison of binary and burst spiking mechanisms. (a) Binary spiking exhibits abrupt state transitions under perturbations, causing error accumulation. (b) Burst spiking smooths transitions between states and reduces perturbation impact. (c) Layer-wise flip rate of the binary-spiking-based Modified PLIF-Net on CIFAR10-C with varying Gaussian noise severities. C1–C17 and Fc represent the convolutional and fully connected layers, respectively. (d) Flip rate comparison across different perturbation levels for different methods, showing that burst spiking enhances robustness.}
  
  \label{fig9}
\end{figure}

\subsection{Burst Spiking Improves Accuracy and Robustness}~\label{bsnbinary}

We propose that burst spiking can enhance both the accuracy and robustness of SNNs. In conventional spiking neurons, the activation state is represented by a single spike and is therefore binary. This simplistic assumption introduces quantization error relative to ANNs, which reduces accuracy~\cite{hu2023fast}. By allowing multiple spikes within a short temporal window, burst spiking alleviates quantization error and strengthens accuracy.

Regarding robustness, in conventional neurons with binary spiking, even small perturbations near the firing threshold can cause abrupt transitions in activation state, leading to accumulated errors across layers and compromised robustness.
Fig.~\ref{fig9_a} and Fig.~\ref{fig9_b} illustrate this intuition. Burst spiking introduces graded activations that smooth transitions between states and mitigate the impact of small perturbations.
To quantitatively assess this effect, we measure the neuron state flip rate under perturbations, where the active state corresponds to positive output and the inactive state to zero output.
As shown in Fig.~\ref{fig9_c}, the flip rate in binary spiking networks increases markedly across layers as the severity of Gaussian noise grows, indicating a clear correlation between flip dynamics and perturbation sensitivity. In this figure, the x-axis represents the network layers from the first convolutional layer (denoted C1) to the fully connected layer (Fc). The five curves (L1 to L5) indicate different levels of perturbation severity.

Building on this flip-based characterization, Fig.~\ref{fig9_d} compares the statistical mean and variance of flip rates across all perturbation severity levels for ANNs, binary spiking networks, and the proposed burst spiking networks.
The results show that the ANN model exhibits substantially lower flip rates than binary spiking, while the proposed burst spiking mechanism significantly narrows this gap, thereby enhancing robustness against perturbations.

\begin{figure}[t]
  \centering

  \begin{subfigure}{\linewidth}
    \centering
    \includegraphics[width=0.9\linewidth]{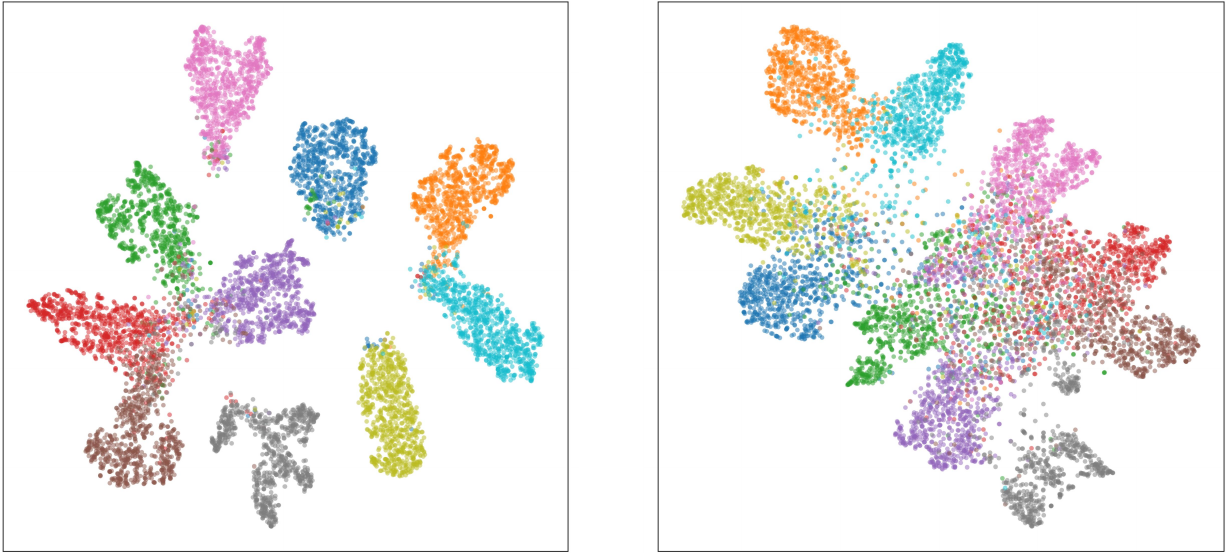}
    \caption{ANN}
    \label{fig:tsne_ann}
  \end{subfigure}

  \begin{subfigure}{\linewidth}
    \centering
    \includegraphics[width=0.9\linewidth]{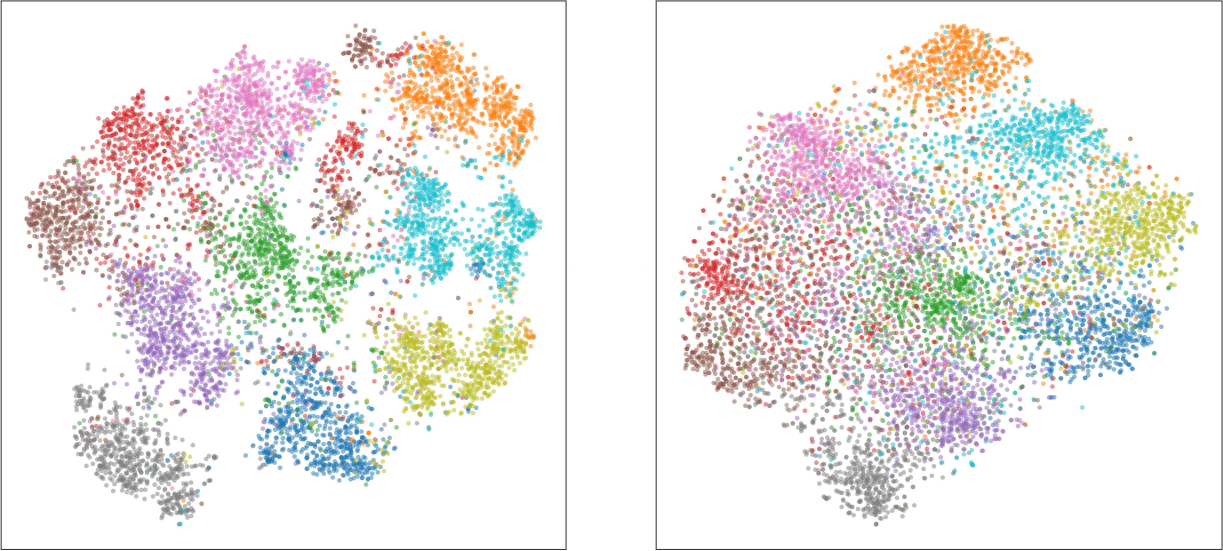}
    \caption{Binary spiking}
    \label{fig:tsne_binary}
  \end{subfigure}

  \begin{subfigure}{\linewidth}
    \centering
    \includegraphics[width=0.9\linewidth]{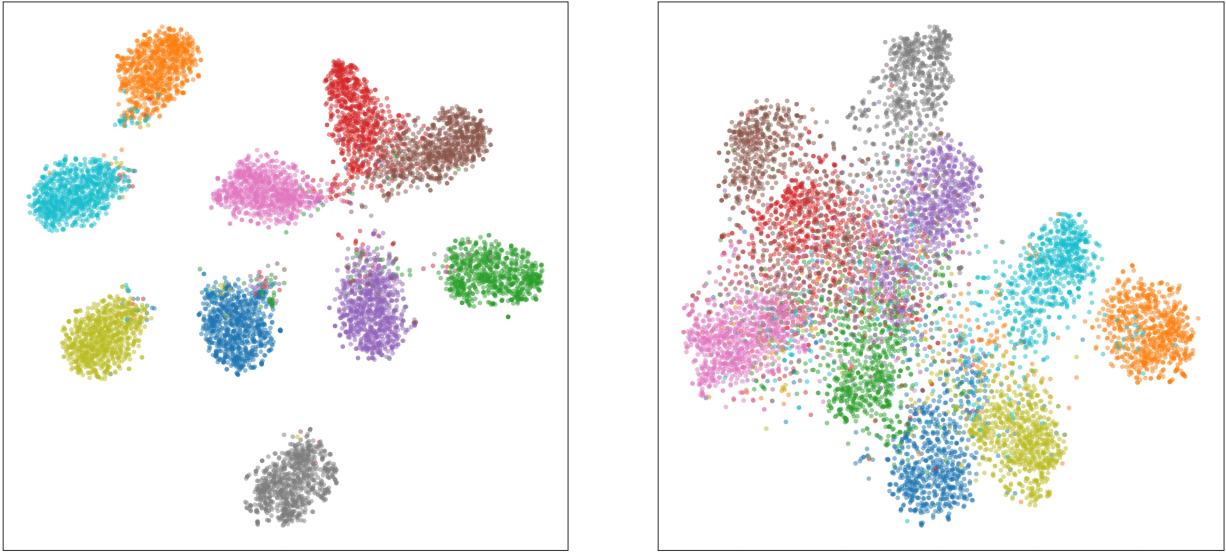}
    \caption{Burst spiking}
    \label{fig:tsne_burst}
  \end{subfigure}

    \caption{t-SNE visualizations of classification-layer features from different models based on the Modified PLIF-Net, using 500 random samples from the CIFAR10-C dataset. Each subfigure shows results under clean inputs (left) and Gaussian noise perturbations (right).}
  \label{fig:tsne}
\end{figure}

Additionally, we visualize the effect of burst spiking on the network’s feature space using t-SNE, as shown in Fig.~\ref{fig:tsne}. 
Compared with the binary spiking model, burst spiking exhibits markedly improved cluster compactness and inter-class separability, even surpassing ANN in clustering quality. 
These results indicate that burst spiking yields more discriminative and stable feature representations, thereby improving both classification accuracy and robustness to input perturbations.

\begin{figure*}[t]
    \centering
    \begin{subfigure}[t]{0.24\textwidth}
        \centering
        \includegraphics[width=\linewidth]{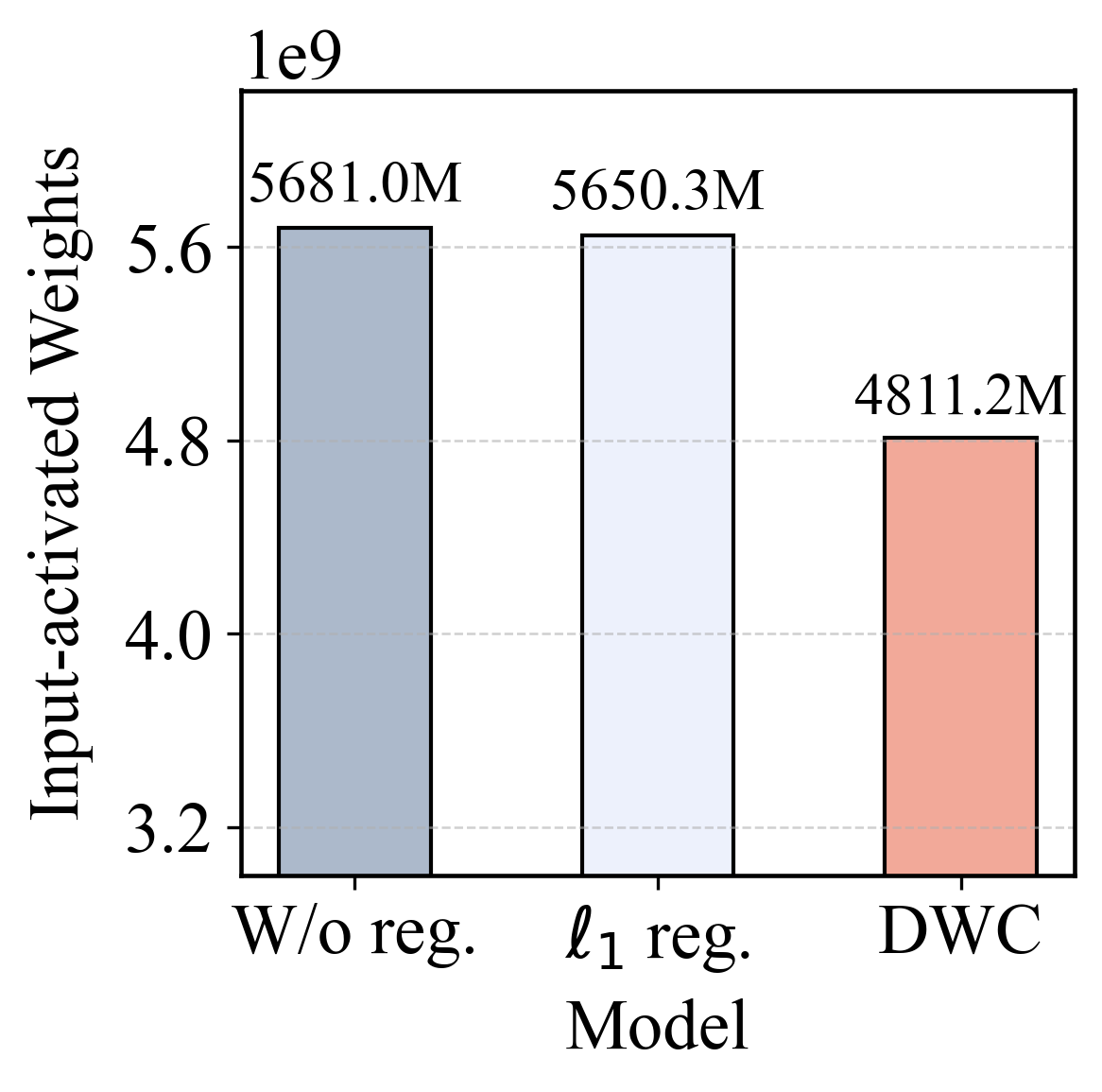}
        \caption{Total}
        \label{fig:activation_heatmap}
    \end{subfigure}
    \begin{subfigure}[t]{0.24\textwidth}
        \centering
        \includegraphics[width=\linewidth]{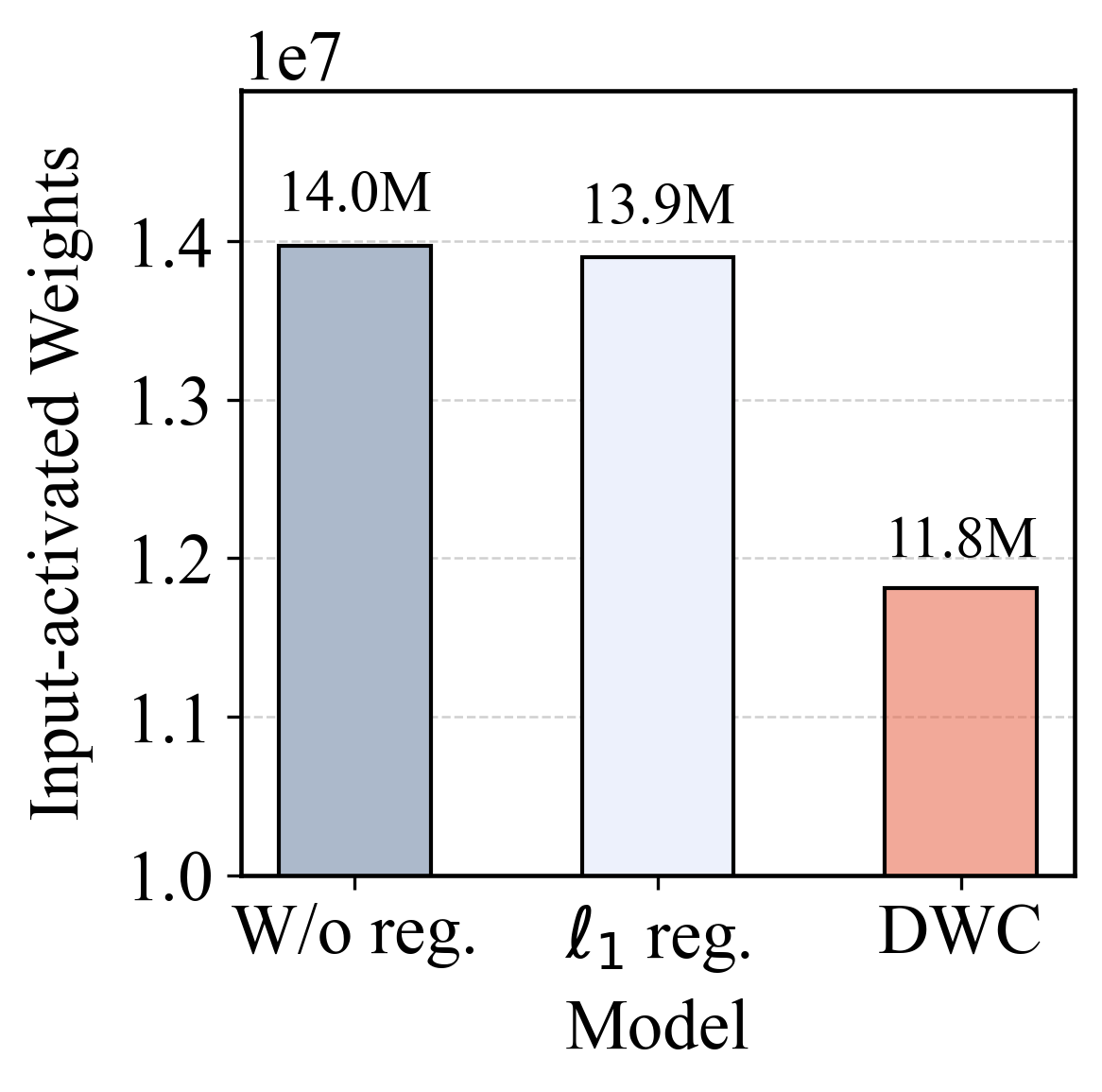}
        \caption{C1}
        \label{fig:dynamic_weights1}
    \end{subfigure}
    \begin{subfigure}[t]{0.24\textwidth}
        \centering
        \includegraphics[width=\linewidth]{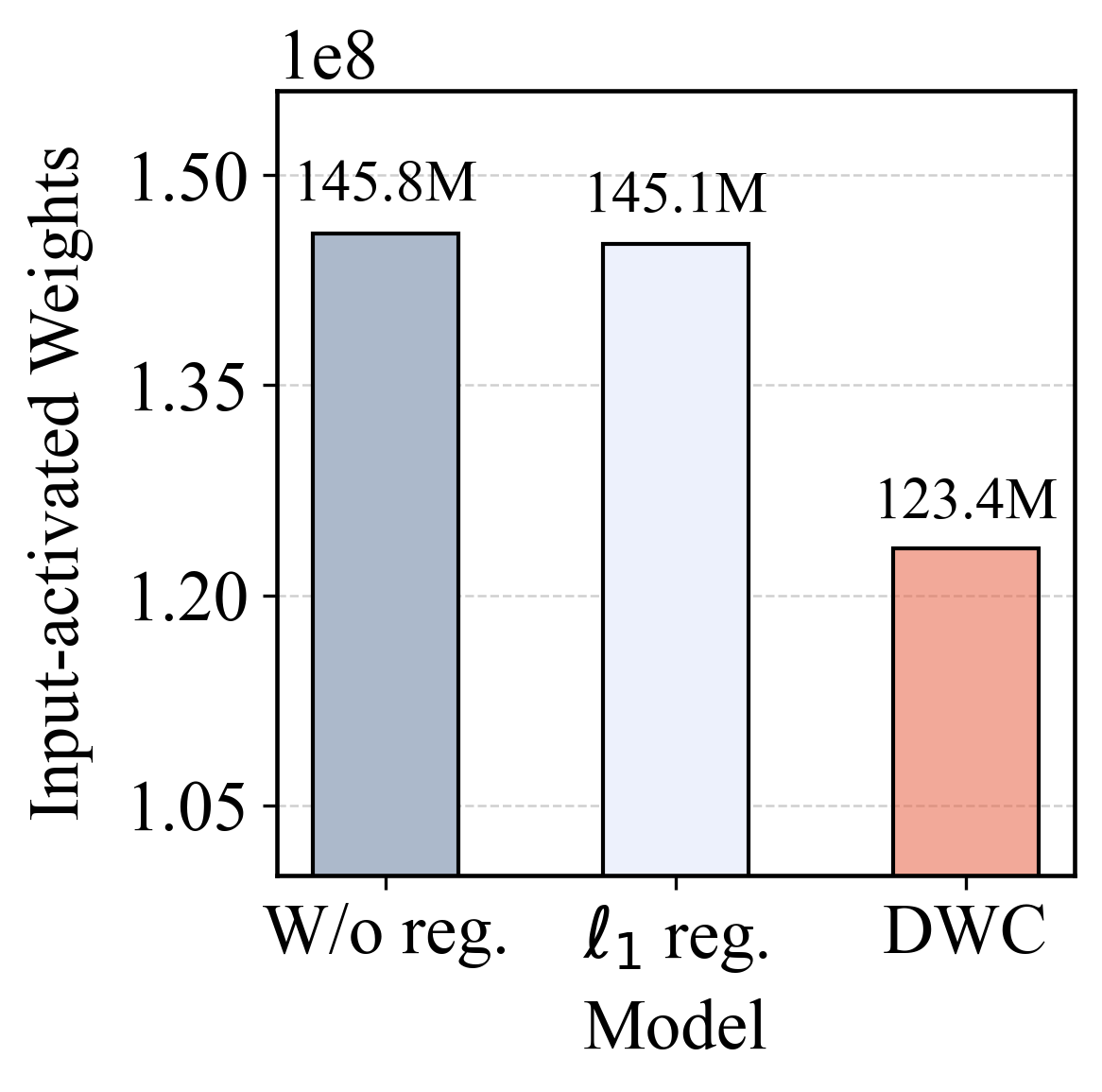}
        \caption{C4}
        \label{fig:dynamic_weights2}
    \end{subfigure}
    \begin{subfigure}[t]{0.24\textwidth}
        \centering
        \includegraphics[width=\linewidth]{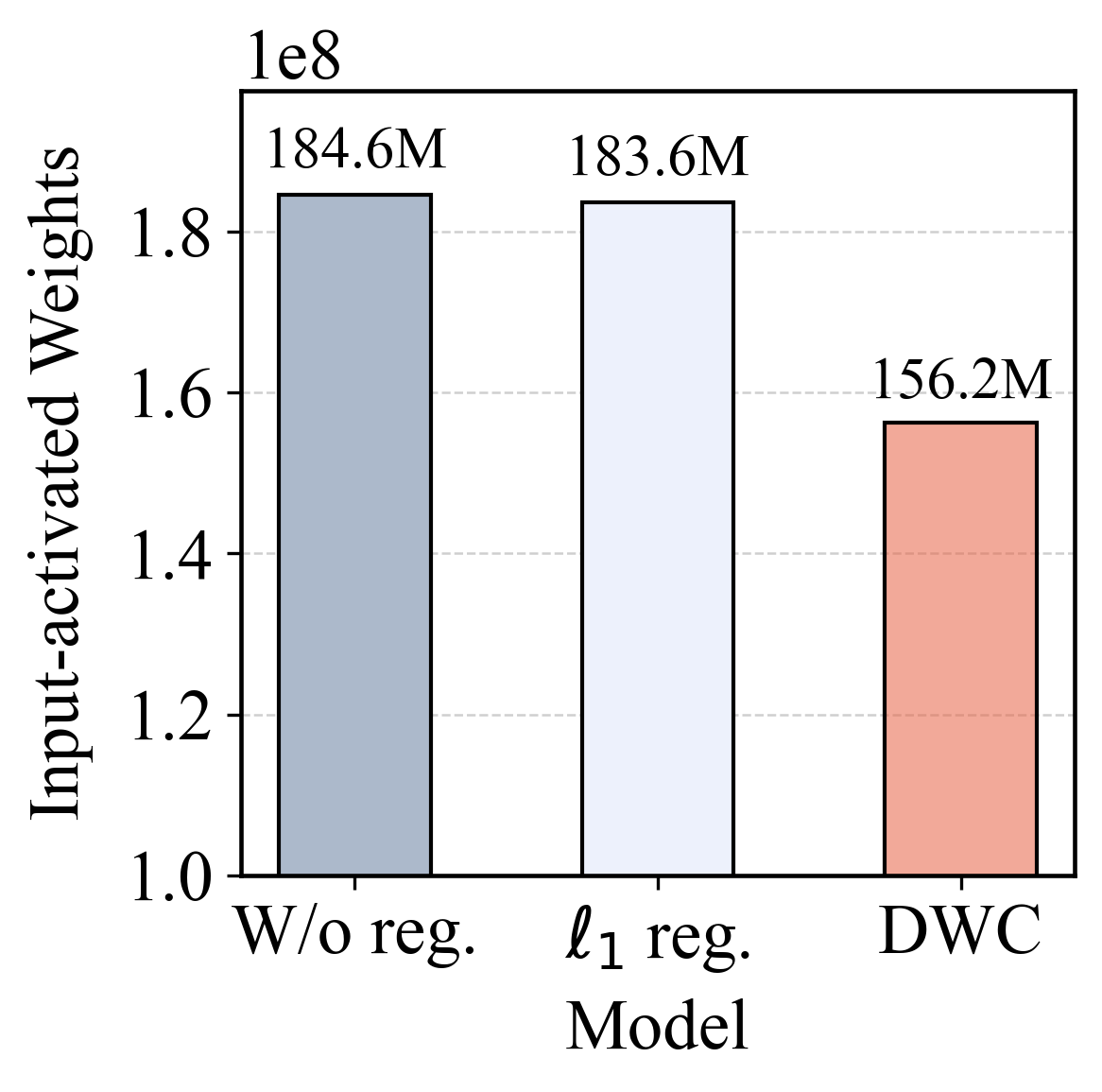}
        \caption{C7}
        \label{fig:dynamic_weights3}
    \end{subfigure}
    \begin{subfigure}[t]{0.24\textwidth}
        \centering
        \includegraphics[width=\linewidth]{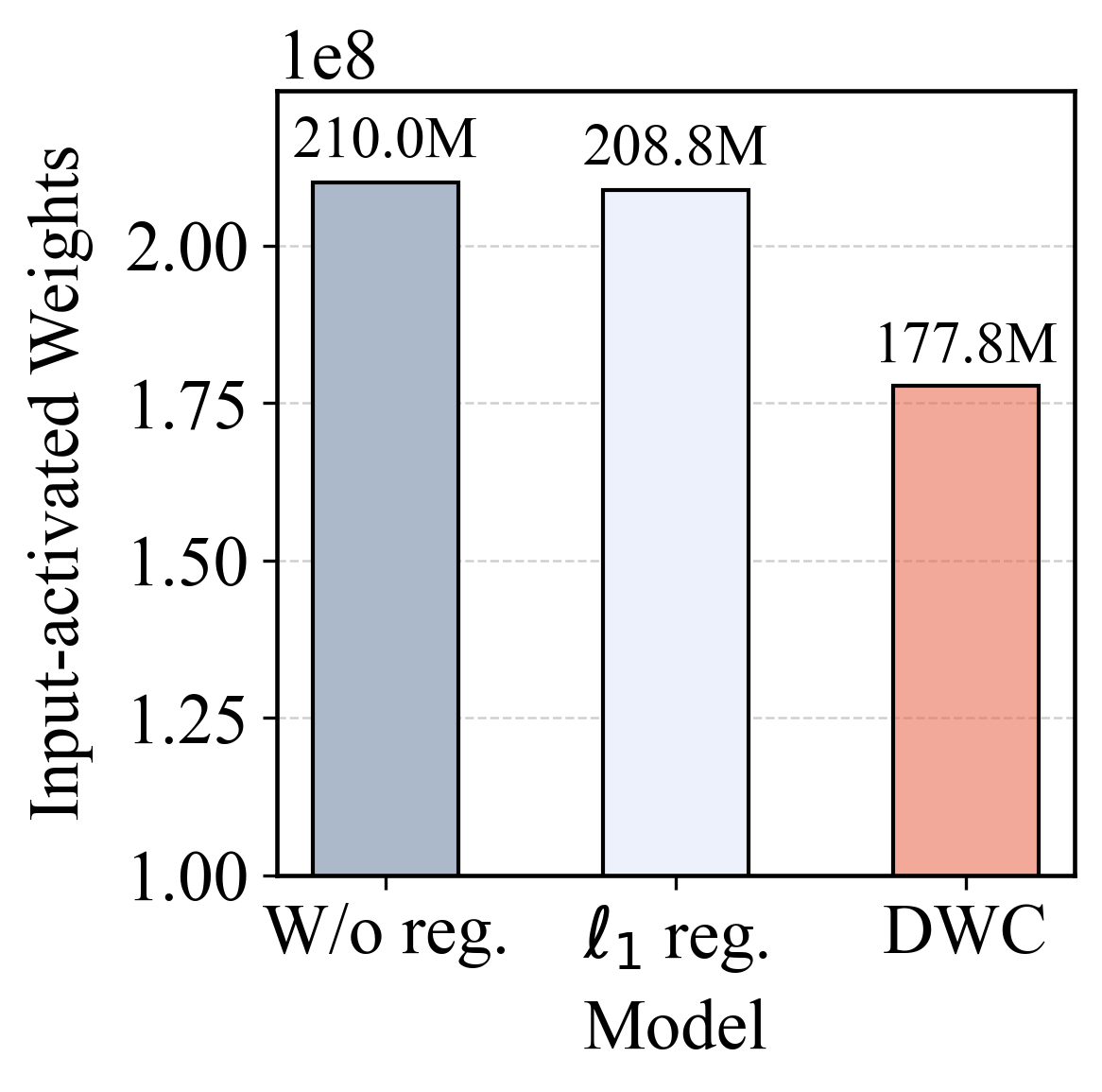}
        \caption{C11}
        \label{fig:dynamic_weights4}
    \end{subfigure}
    \begin{subfigure}[t]{0.24\textwidth}
        \centering
        \includegraphics[width=\linewidth]{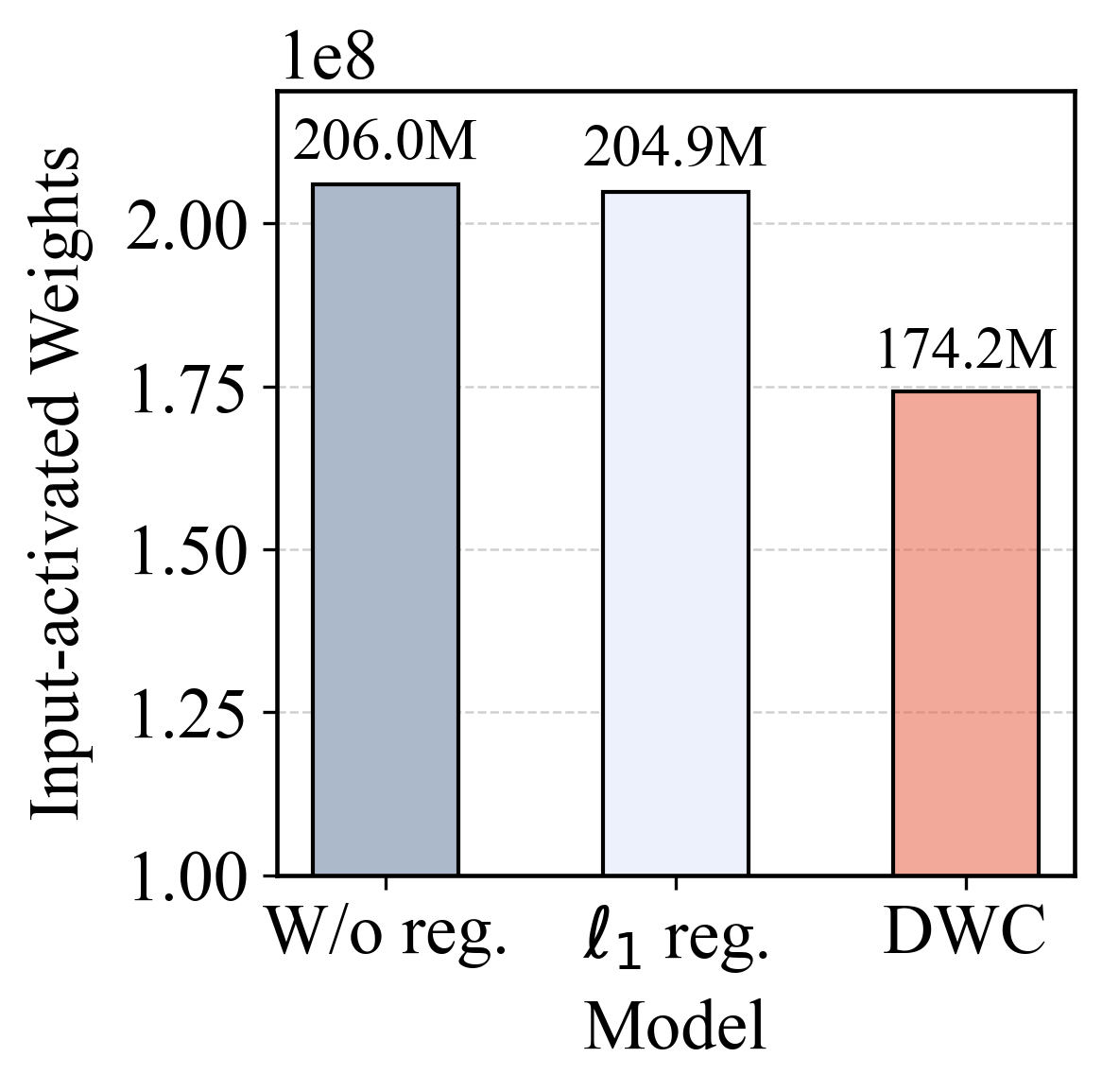}
        \caption{C14}
        \label{fig:dynamic_weights5}
    \end{subfigure}
    \begin{subfigure}[t]{0.24\textwidth}
        \centering
        \includegraphics[width=\linewidth]{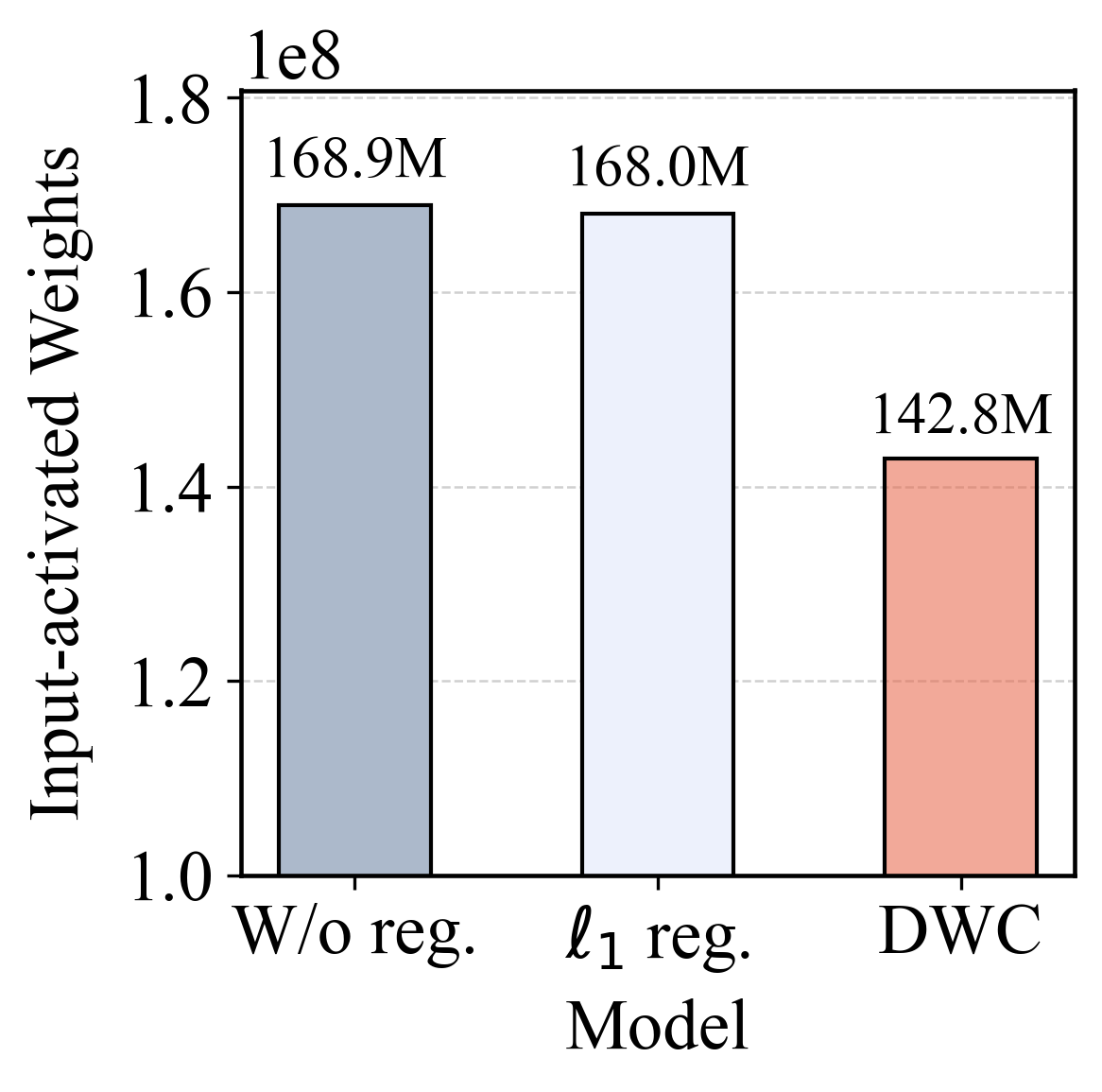}
        \caption{C17}
        \label{fig:dynamic_weights6}
    \end{subfigure}
    \begin{subfigure}[t]{0.24\textwidth}
        \centering
        \includegraphics[width=\linewidth]{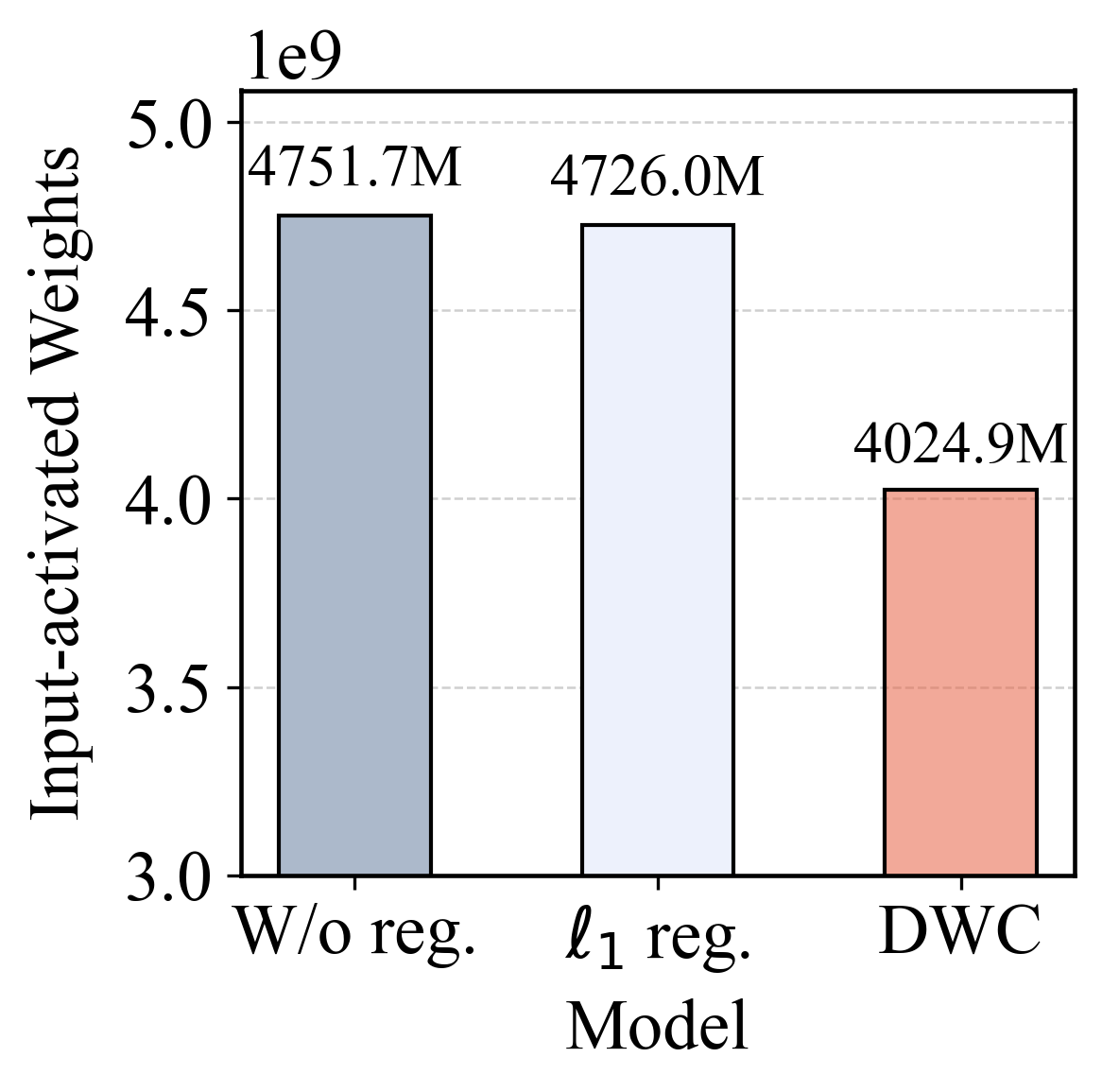}
        \caption{Fc}
        \label{fig:dynamic_weights7}
    \end{subfigure}

    \caption{Comparison of input-activated weights of Modified PLIF-Net on the CIFAR-10 dataset.
(a) Overall number of input-activated weights across different methods.
(b–h) Layer-wise number of input-activated weights under various regularization strategies. C1–C17 and Fc represent the convolutional and fully connected layers, respectively.
}
    \label{fig:active_weight_norm}
\end{figure*}

\subsection{DWC Reduces Weights Involved in Computation} \label{exp:DWC}

\rev{Experimental results demonstrate that DWC contributes to improving both the accuracy and robustness of BuSNNs. The key motivation behind this improvement is that DWC effectively constrains connection weights, according to Eq.~\ref{eq:weight_amplification_bound}. }
We specifically analyze the input-activated connections, whose presynaptic neurons are active during a forward pass. Only these connections participate in the actual computation and influence the output for a given input, with a small magnitude reflecting lower sensitivity of the model to input perturbations. Specifically, we define the input-activated weights as the sum of the absolute values of the weights on these connections, which reflects the total strength of the connections actively involved in the computation.

We analyze the input-activated weights obtained from the BuSNN using the Modified PLIF-Net architecture under three different training strategies: a baseline without regularization, $\ell_1$ regularization, and DWC.
The input-activated weights are computed across the CIFAR-10 dataset, considering all input samples. This allows us to assess the total number of weights involved in the model's computations across different inputs, providing a comprehensive view of the weights actively contributing to the model's output. As illustrated in Fig.~\ref{fig:active_weight_norm}, the input-activated weights of the models trained without regularization and with $\ell_1$ regularization are nearly identical, indicating that conventional $\ell_1$ regularization has a limited effect in reducing the weights that actively participate in computation.
In contrast, the DWC strategy yields a substantial reduction in input-activated weights. 
These results suggest that DWC effectively suppresses the magnitude of weights actively involved in computation, thereby reducing the model’s sensitivity to input perturbations. 


\section{Conclusion and Discussion}

In this work, we investigate the performance of SNNs in terms of accuracy on clean data and robustness under adversarial attacks and common corruptions. Through controlled comparisons with ANNs using identical architectures and training settings, we reveal that conventional SNNs still exhibit limited performance in both aspects. To address these issues, we propose BuSNNs, which consist of BSNs and DWC. BSNs enhance representational capacity and perturbation robustness by using burst spiking. DWC further improves performance by effectively constraining weight magnitudes based on the activation state of BSNs. Extensive experiments confirm that our method consistently improve both accuracy and robustness. These findings provide new insights into the design of robust and efficient SNNs and highlight the importance of jointly optimizing activations and weights in future SNN development.

Our analysis of neural activation offers a new perspective on the role of the multi-spike mechanism in SNNs. Previous works~\cite{ponghiran2022spiking,luo2024integer,shen2024conventional} have investigated this mechanism primarily from the standpoint of reducing quantization error, showing that multi-spike neurons can achieve higher accuracy than binary ones. SFA~\cite{yao2025scaling} further demonstrates that multi-spike firing can be feasibly implemented on neuromorphic hardware, providing a theoretical advantage in energy efficiency. \rev{Unlike prior multi-spike designs that mainly aim to reduce binary quantization error, BSN formulates burst firing as a membrane-potential-driven response. The suprathreshold potential strength is reflected by a bounded burst count controlled by $\theta_{\text{burst}}$. Our results further reveal that this multi-spike mechanism not only improves accuracy but also enhances robustness. By allowing each neuron to represent graded activation strength within a short temporal window, BSNs provide finer-grained neuronal responses and reduce the quantization error caused by binary firing. Meanwhile, the graded burst levels smooth the transition between activation states, thereby improving robustness.}

In addition, effectively training high-performance SNNs has long been a central challenge in the field. The surrogate gradient method, which allows SNNs to be optimized directly toward target objectives in a manner similar to ANNs, has become one of the mainstream approaches. Motivated by our focus on robustness, we introduce a weight optimization mechanism DWC inspired by biological synaptic plasticity. \rev{Such activation-dependent weight control reduces the amplification of input perturbations through \rev{network} layers, thereby limiting output variation and emphasizing the crucial role of biologically grounded constraints in SNN training.} 
We therefore argue that SNNs differ from ANNs in their training dynamics and connectivity constraints. 
Therefore, developing superior SNN models may require deeper biological inspiration to better exploit the unique computational properties of SNNs.

\rev{In our experiments, the proposed models {outperform matched ANN counterparts} in both accuracy and robustness on small-scale benchmarks, such as CIFAR-10 with Modified PLIF-Net and CIFAR-100 with VGG11. On large-scale benchmarks, including ImageNet, {BuSNN consistently improves over existing SNNs}. In matched IMAGENET-C settings, BuSNN surpasses INT4 and INT8 activation-quantized ANN baselines for ResNet-18 and ResNet-34, while also outperforming the INT4 baseline for ResNet-50. However, {full-precision ANN baselines still retain stronger performance in most large-scale settings}. {Further evaluation on larger architectures, stronger ANN training recipes, and broader perception tasks remains an important direction for future work.}}

\rev{Since BSNs act on spike-based activation representation and DWC functions as an activation-dependent weight constraint, they can be incorporated into different SNN training objectives without changing the overall learning framework. For instance, the proposed mechanisms are potentially compatible with advanced self-supervised learning~\cite{yang2025selfsupervised} and other training strategies~\cite{yang2023sibols, shen2026stage} for improving SNN representation and robustness. This compatibility provides a practical direction for extending BuSNNs to broader training paradigms and perception tasks.} These results demonstrate that SNNs hold strong potential for lightweight, energy-efficient, and robust perception tasks, offering a promising alternative to conventional ANNs in practical scenarios.

\section{Limitations}

\rev{
This study still has several limitations. First, we do not include comparisons with state-of-the-art ANN models that rely on highly specialized architectures, large-scale pretraining, or advanced data augmentation. Instead, we focus on standard ANN baselines under matched architectures and training protocols. This setting enables a fair and interpretable comparison between SNNs and ANNs, since the influence of the computational paradigm can be better isolated. Many advanced ANN techniques, such as data augmentation and knowledge distillation, are also complementary to our framework and can potentially be integrated into future SNN training pipelines.}
\rev{Second, BuSNN does not fully close the gap to ANNs on large-scale datasets. Full-precision ANNs still retain stronger performance under larger backbones.}
\rev{Third, BuSNN has mainly been evaluated in classification-oriented settings. Its generalizability to broader vision tasks, such as object detection and semantic segmentation, remains to be explored. Future work will further address these limitations.}


\section{\rev{Data Availability}}

\rev{All datasets used in this study are publicly available open-source datasets. The code will be publicly available on \url{https://github.com/fourbeans/BuSNN}.}

\bibliographystyle{ieeetr}
\bibliography{ref}


\section*{Biography Section}
\vspace{-3em}
\begin{IEEEbiography}
[{\includegraphics[width=1in,height=1.25in,clip,keepaspectratio]{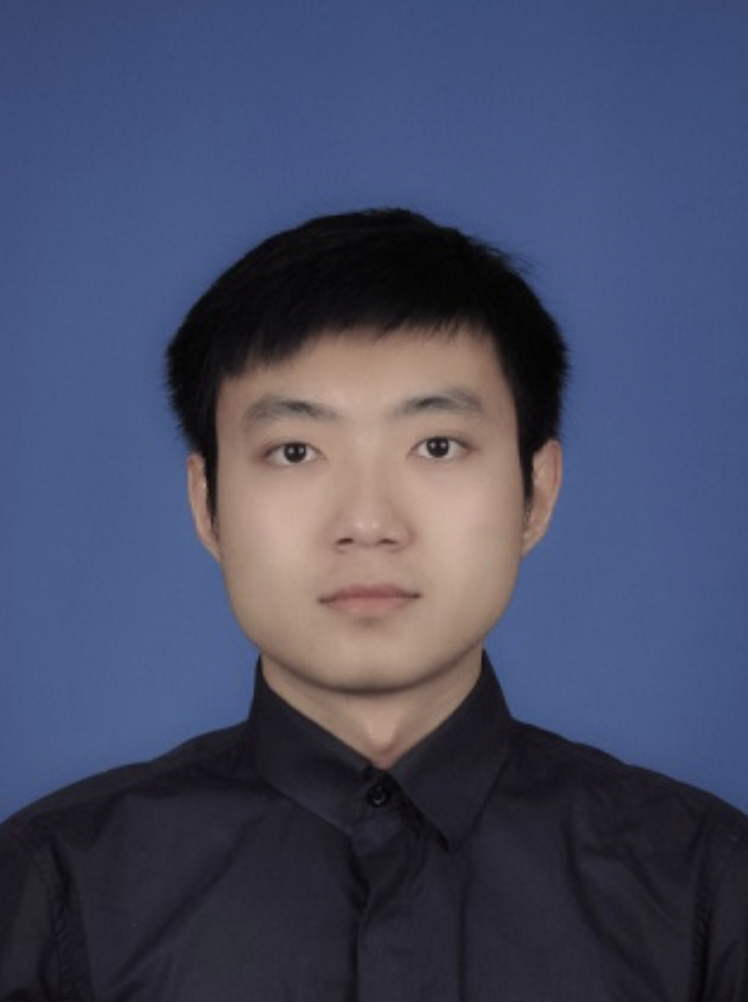}}]{Jiahong Zhang}
received the B.S. degree from Sichuan University, Chengdu, China, in 2019, and the M.S. degree from the State Key Laboratory of Media Convergence and Communication, Communication University of China, Beijing, China, in 2023. He is currently pursuing the Ph.D. degree with the Institute of Automation, Chinese Academy of Sciences (CASIA), Beijing.
His research interests include computer vision and brain-inspired intelligence. 
\end{IEEEbiography}

\vspace{-4em}
\begin{IEEEbiography}[{\includegraphics[width=1in,height=1.25in,clip,keepaspectratio]{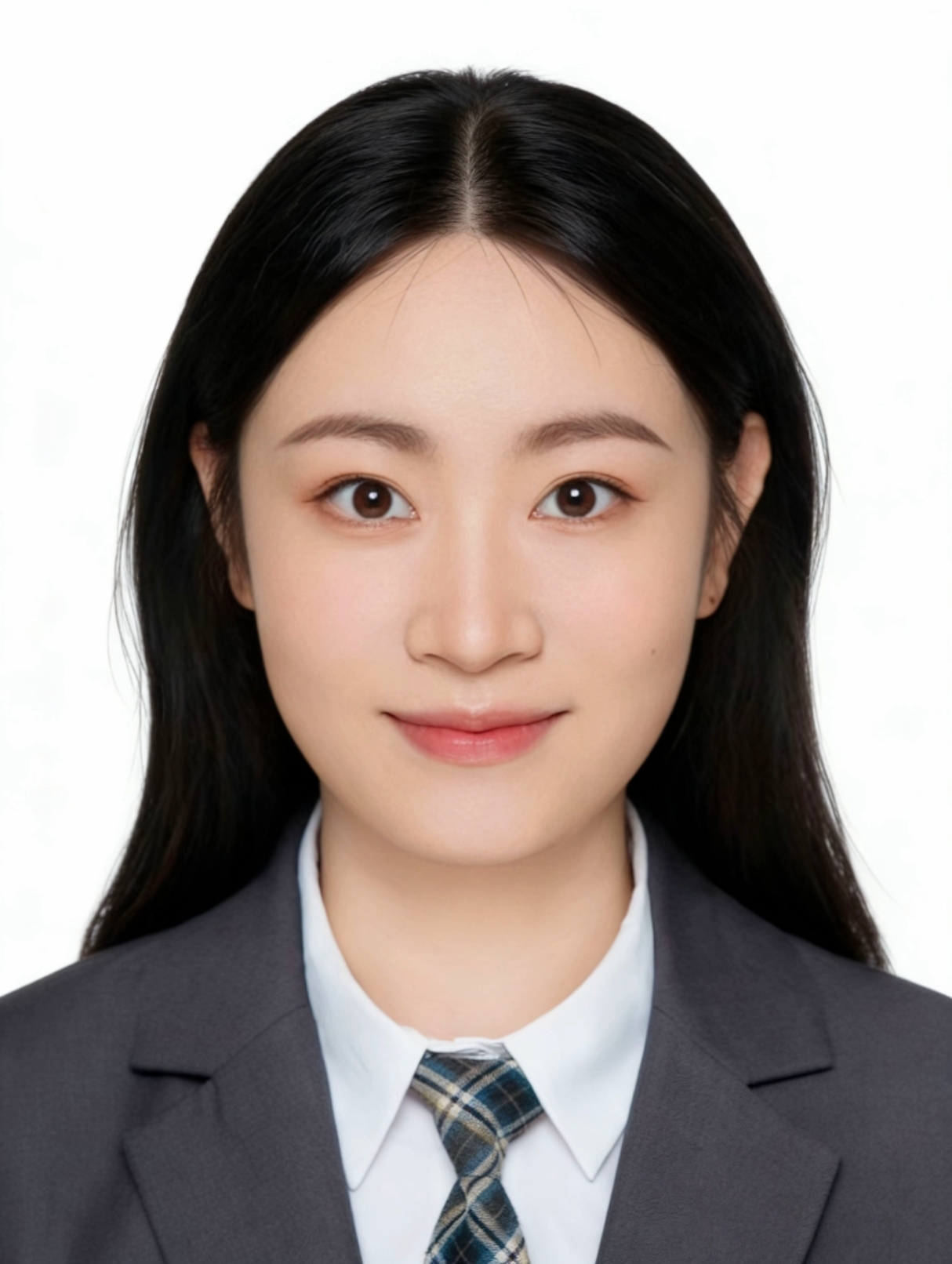}}]{Sijun Shen}
is currently a joint Master’s student at the State Key Laboratory of Media Convergence and Communication, Communication University of China, Beijing. Her research focuses on neural computation and spiking neural networks, with a focus on developing low-power neuromorphic computing and brain-inspired learning algorithms, and demonstrating their effectiveness in solving real-world problems.
\end{IEEEbiography}

\vspace{-3em}
\begin{IEEEbiography}[{\includegraphics[width=1in,height=1.25in,clip,keepaspectratio]{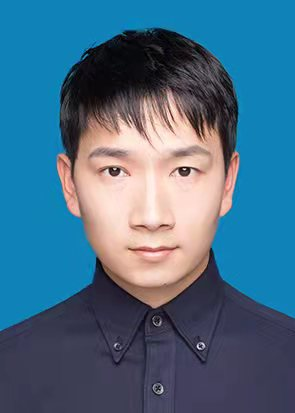}}]{Man Yao}
received the Ph.D. from Xi’an Jiaotong University, Xi’an, China, in 2023. He is currently an Assistant Professor at the Institute of Automation, Chinese Academy of Science (CASIA). He has published several papers in journals, including Nature Communications, IEEE T-PAMI, and Neural Networks, as well as at AI conferences such as ICLR, NeurIPS, ICML, and ICCV, where he served as the first author. His research interests include neuromorphic computing and dynamic neural networks.
\end{IEEEbiography}

\vspace{-3em}
\begin{IEEEbiography}[{\includegraphics[width=1in,height=1.25in,clip,keepaspectratio]{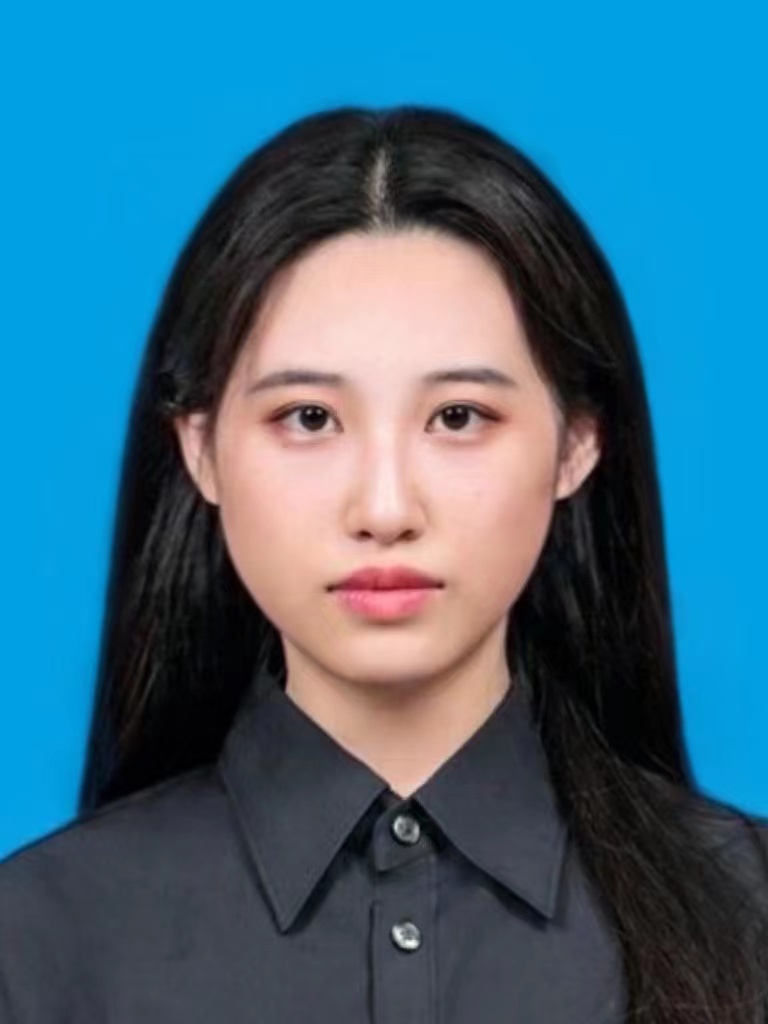}}]{Han Xu}
is currently a joint Ph.D. student at the Institute of Automation, Chinese Academy of Sciences and the Beijing Academy of Artificial Intelligence. The research focuses on brain-inspired large models, with particular emphasis on integrating neuroscience mechanisms with artificial intelligence technologies to develop next-generation general intelligence models.
\end{IEEEbiography}

\vspace{-3em}
\begin{IEEEbiography}[{\includegraphics[width=1in,height=1.25in,clip,keepaspectratio]{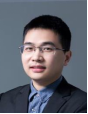}}]{Mingqiang Huang}
received the B.S. and Ph.D. degrees from Huazhong University of Science and Technology, Wuhan, China, in 2013 and 2018, respectively. Then he became a Research Fellow in Nanyang Technological University, Singapore. From 2019 to 2026, he works as a Research Associate Professor in Shenzhen Institute of Advanced Technology, Chinese Academy of Sciences. Currently he is with Wuhan University, China. His research interests include energy-efficient computing, artificial intelligence (AI) hardware accelerators.
\end{IEEEbiography}

\vspace{-2em}
\begin{IEEEbiography}[{\includegraphics[width=1in,height=1.25in,clip,keepaspectratio]{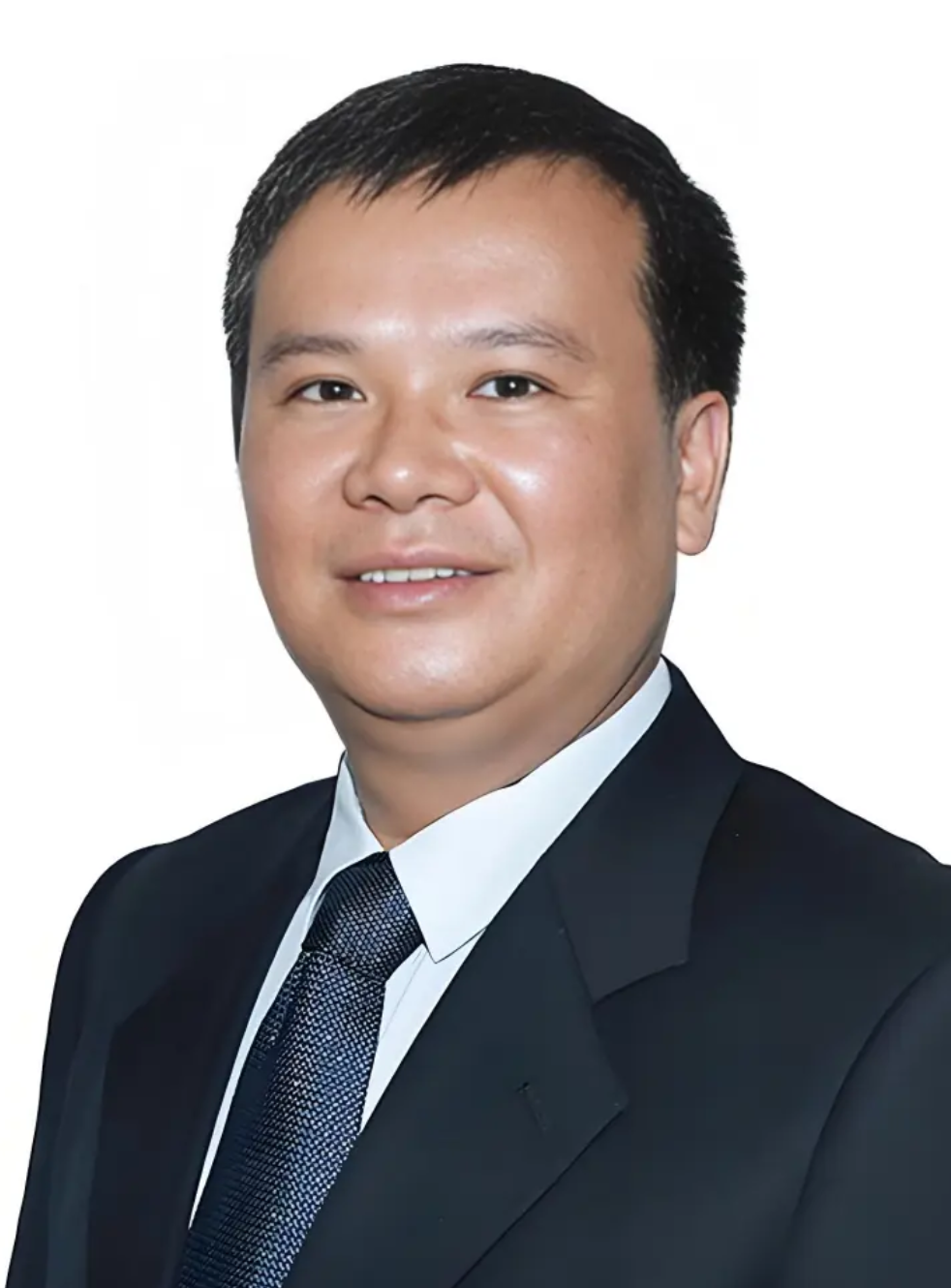}}]{Yonghong Tian}
(Fellow, IEEE) is currently a boya distinguished professor with the School of Computer Science, Peking University, China, and is also the deputy director of Artificial Intelligence Research Center, PengCheng Laboratory, Shenzhen, China. 
His research interests include neuromorphic vision, distributed machine learning and multimedia Big Data. 
He is the author or coauthor of more than 280 technical articles in refereed journals and conferences.
He was/is an associate editor of IEEE Transactions on Circuits and Systems for Video Technology (2018.1- 2021.12), IEEE Transactions on Multimedia (2014.8-2018.8), IEEE Multimedia Magazine (2018.1-), and IEEE Access (2017.1-). He co-initiated International Conference on Multimedia Big Data (BigMM) and served as the TPC Co-chair of BigMM 2015, and aslo served as the Technical Program Co-chair of IEEE ICME 2015, IEEE ISM 2015 and IEEE MIPR 2018/2019, and General co-chair of IEEE MIPR 2020 and ICME2021. He is the steering member of IEEE ICME (2018-2020) and IEEE BigMM (2015-), and is a TPC Member of more than ten conferences such as CVPR, ICCV, ACM KDD, AAAI, ACM MM and ECCV. He was the recipient of the Chinese National Science Foundation for Distinguished Young Scholars in 2018, two National Science and Technology Awards and three ministerial-level awards in China, and obtained the 2015 EURASIP Best Paper Award for Journal on Image and Video Processing, and the best paper award of IEEE BigMM 2018. He is a senior member of ICIE and CCF, a member of ACM.
\end{IEEEbiography}

\vspace{-2em}
\begin{IEEEbiography}[{\includegraphics[width=1in,height=1.25in,clip,keepaspectratio]{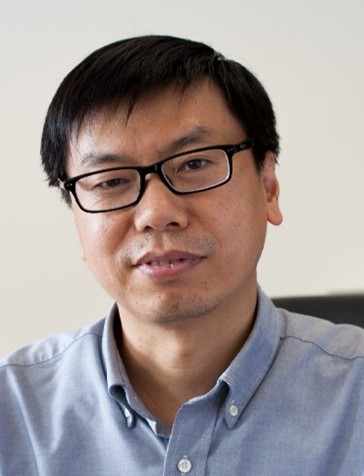}}]{Bo Xu}
received the BS degree from Zhejiang University, Hangzhou, China, in 1988, and the MS and PhD degrees from the University of Chinese Academy of Sciences, Beijing, China, in 1992 and 1997, respectively. Currently, he is a professor with the Institute of Automation, Chinese Academy of Science(CASIA). And now is the director of CASIA, associate director of CAS(Chinese Academy of Sciences) Center for Excellence in Brain Science and Intelligence Technology(CEBSIT). His research interests include spans the area of speech recognition, spoken dialogue, multimodality information processing, brain-inspired learning and decision intelligence. He once was the chairman of Special Interesting Group in International Association of Chinese Spoken language Processing(SIG-ISCSLP) between 2001-2006, and the Coordinator of International Consortium for Speech Translation Advanced Research (C-Star). 
He was steering committee member of National High-Tech Program in the fields of intelligence information processing and interactive interfaces from 2004-2010. 
He won several prizes including the best paper of ISCSLP, outstanding young scientist of CAS and outstanding achievement in national media technology development. He has published more than 300 papers and holds 50 patents in the fields. He now serves as steering committee member of National Programme of New Generation Artificial Intelligence from 2017.
\end{IEEEbiography}

\vspace{-2em}
\begin{IEEEbiography}[{\includegraphics[width=1in,height=1.25in,clip,keepaspectratio]{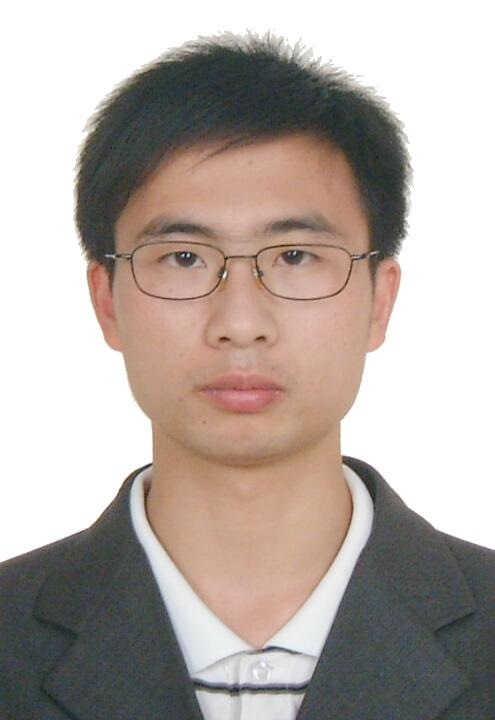}}]{Guoqi Li}
received the PhD degree from Nanyang Technological University, Singapore, in 2011. From 2011 to 2014, he was a scientist with the Data Storage Institute and the Institute of High Performance Computing, Agency for Science, Technology and Research, Singapore. From 2014–2022, he was an assistant professor and associate professor with Tsinghua University, Beijing, China. Since 2022, he has been with the Institute of Automation,Chinese Academy of Sciences and the University of Chinese Academy of Sciences, where he is currently a full professor. 
His current research interests include Brain-inspired Intelligence, Neuromorphic Computing and Spiking Neural Networks.
 He has authored or co-authored more than 170 papers in a number of prestigious journals including Nature, Nature Communications, Science Robotics, Proceedings of the IEEE, and top AI conference such as ICLR, NeurIPS, ICML, AAAI and so on. He has been actively involved in professional services such as serving as a Tutorial Chair, an International Technical Program Committee Member, a PC member, a Publication Chair, a Track Chair and workshop chair for several international conferences. He is an Editorial-Board Member for Control and Decision, and served as associate editors for Journal of Control and Decision and Frontiers in Neuroscience: Neuromorphic Engineering. He is a reviewer for Mathematical Reviews published by the American Mathematical Society and serves as a reviewer for a number of prestigious international journals and top AI conferences including ICLR, NeurIPS, ICML, AAAI and so on.
He was the recipient of the 2018 First Class Prize in Science and Technology of the Chinese Institute of Command and Control, the Top ten scientific advances Award in China selected by the Ministry of science and technology, P.R. China as the backbone of the team member, and the 2020 Second Prize of Fujian Provincial Science and Technology Progress Award. He received the outstanding Young Talent Award of the Beijing Natural Science Foundation in 2021.
\end{IEEEbiography}

\end{document}